\documentclass[lettersize,journal]{IEEEtran}
\usepackage{amsmath,amsfonts}
\usepackage{algorithmic}
\usepackage{algorithm}
\usepackage{array}
\usepackage[caption=false,font=normalsize,labelfont=sf,textfont=sf]{subfig}
\usepackage{booktabs}
\usepackage{textcomp}
\usepackage{stfloats}
\usepackage{url}
\usepackage{verbatim}
\usepackage{graphicx}
\usepackage{cite}
\usepackage{amssymb}
\usepackage{multirow}
\usepackage{color,xcolor}
\usepackage{bbm}
\usepackage{hyperref}
\usepackage{threeparttable}
\usepackage{multirow}
\usepackage{amsmath}
\usepackage{enumerate}

\makeatletter
\renewcommand{\maketag@@@}[1]{\hbox{\m@th\normalsize\normalfont#1}}%
\makeatother

\begin{document}
\title{DA$^{2}$Diff: Exploring Degradation-aware Adaptive Diffusion Priors for All-in-One Weather Restoration}

\author{Jiamei Xiong, Xuefeng Yan, Yongzhen Wang, Wei Zhao, Xiao-Ping Zhang, and Mingqiang Wei

\thanks{Jiamei Xiong, Wei Zhao and Mingqiang Wei are with the School of Computer Science and Technology, Nanjing University of Aeronautics and Astronautics, Nanjing 210016, China (e-mail: jmxiong@nuaa.edu.cn, weizhao0120@nuaa.edu.cn, mingqiang.wei@gmail.com).}
\thanks{Xuefeng Yan is with the School of Computer Science and Technology, Nanjing University of Aeronautics and Astronautics, Nanjing 210016, and also with the Collaborative Innovation Center of Novel Software Technology and Industrialization, Nanjing 210093, China (e-mail: yxf@nuaa.edu.cn).}
\thanks{Yongzhen Wang is with the College of Computer Science and Technology, Anhui University of Technology, Ma’anshan 243099, China (e-mail: wangyz@ahut.edu.cn).}
\thanks{Xiao-Ping Zhang is with the Tsinghua Shenzhen International Graduate School, Tsinghua University, Shenzhen, China. (e-mail: xpzhang@ieee.org).}
}

\markboth{Journal of \LaTeX\ Class Files,~Vol.~14, No.~8, August~2021}%
{Shell \MakeLowercase{\textit{et al}}: A Sample Article Using IEEEtran.cls for IEEE Journals}

\maketitle
\begin{abstract}
Image restoration under adverse weather conditions is a critical task for many vision-based applications. Recent all-in-one frameworks that handle multiple weather degradations within a unified model have shown potential. However, the diversity of degradation patterns across different weather conditions, as well as the complex and varied nature of real-world degradations, pose significant challenges for multiple weather removal. To address these challenges, we propose an innovative diffusion paradigm with degradation-aware adaptive priors for all-in-one weather restoration, termed DA$^{2}$Diff. It is a new exploration that applies CLIP to perceive degradation-aware properties for better multi-weather restoration. Specifically, we deploy a set of learnable prompts to capture degradation-aware representations by the prompt-image similarity constraints in the CLIP space. By aligning the snowy/hazy/rainy images with snow/haze/rain prompts, each prompt contributes to different weather degradation characteristics. The learned prompts are then integrated into the diffusion model via the designed weather-specific prompt guidance module, making it possible to restore multiple weather types. To further improve the adaptiveness to complex weather degradations, we propose a dynamic expert selection modulator that employs a dynamic weather-aware router to flexibly dispatch varying numbers of restoration experts for each weather-distorted image, allowing the diffusion model to restore diverse degradations adaptively. Experimental results substantiate the favorable performance of DA$^{2}$Diff over state-of-the-arts in quantitative and qualitative evaluation. Source code will be available after acceptance.
\end{abstract}


\section{Introduction}
\IEEEPARstart{W}{eather} conditions, as common climatic phenomena, inevitably degrade the visibility of images and hamper the performance of downstream vision tasks like object detection \cite{dong2024gmtnet, zhu2023intermediate} and scene understanding \cite{cong2024end, wang2023dcfp}. Therefore, removing weather degradations plays a crucial role in the safety and reliability of outdoor vision systems.

\begin{figure}[t] 
    \centering
    \includegraphics[width=1.0\linewidth]{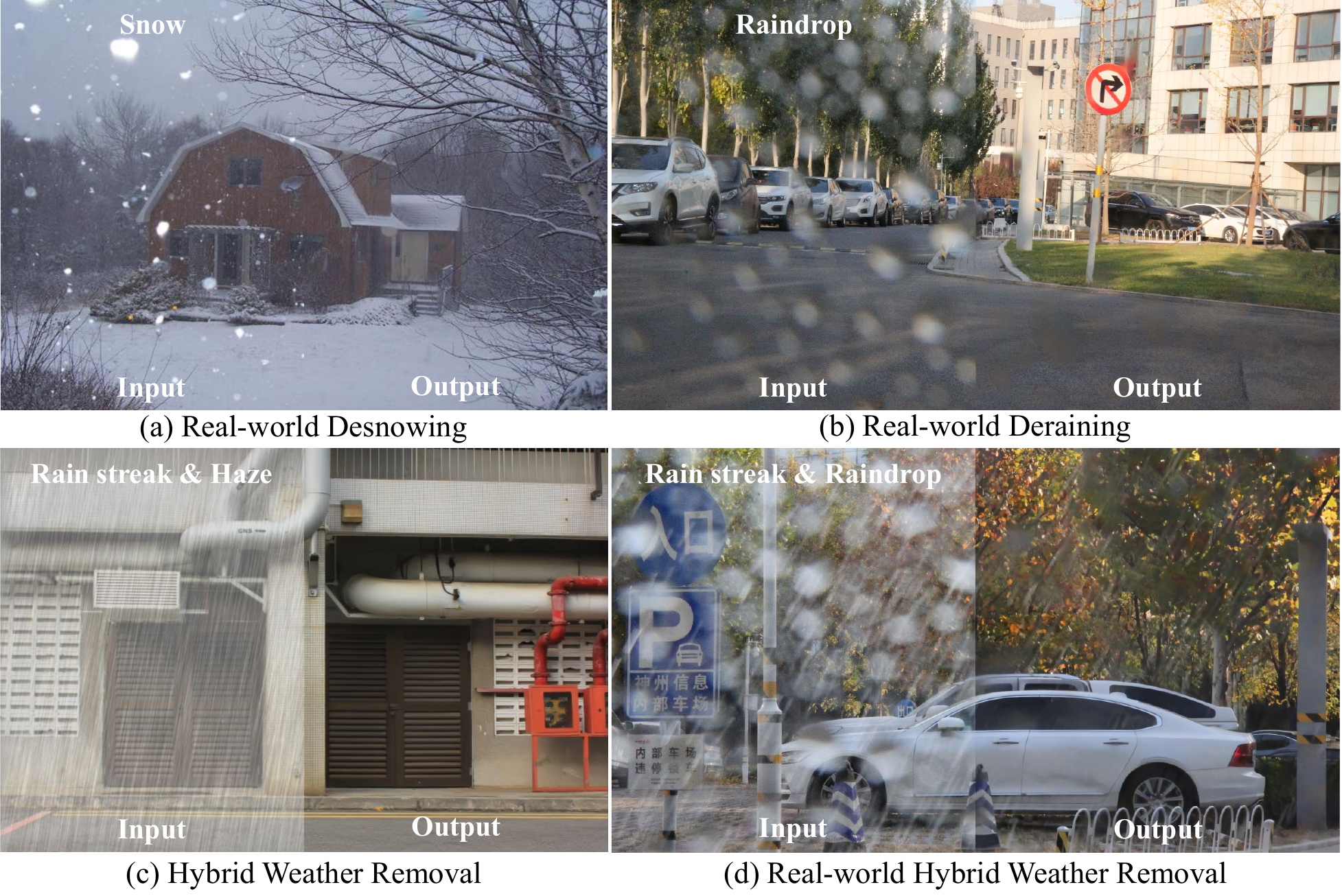}
	\caption{Visual results generated by our DA$^{2}$Diff. Our method is capable of adaptively generating high-fidelity restoration results for the real-world weather degradations.}
    \label{fig: Visual results}
\end{figure}

Learning-based weather restoration methods have achieved remarkable progress. Early efforts focus on restoring specific weather degradation, such as dehazing \cite{he2010single, cai2016dehazenet, liu2019griddehazenet, song2022wsamf, wang2024uncertainty}, deraining \cite{li2018recurrent, qian2018attentive, wang2019spatial, du2020conditional, cai2022multiscale, yan2024glgfn}, and desnowing \cite{liu2018desnownet, chen2020jstasr, jaw2020desnowgan, zhang2021deep}. These methods perform well under specific weather conditions but struggle with others, limiting their practical applicability in scenarios where diverse weather conditions coexist. Since then, various methods \cite{zamir2021multi, chen2022simple, zamir2022restormer, tu2022maxim} have exploited a single model to tackle multiple degradations with task-specific pre-trained weights. Nevertheless, the model requires distinct pre-trained weights for each task, resulting in inflexibility and inefficiency.

Several works \cite{li2020all, chen2022learning, valanarasu2022transweather, zhu2023learning, ozdenizci2023restoring, gao2023frequency, zhu2024mwformer, li2024multi} develop all-in-one models to simultaneously handle multiple weather types with a set of pre-trained weights. Since each weather condition exhibits unique characteristics, restoring them together may cause potential conflicts. To handle the diversity of weather degradations, some approaches design dedicated components for different weather conditions. Specifically, multiple encoders \cite{li2020all} or knowledge learning techniques \cite{chen2022learning} are used to tailor the model for each weather type, but these networks are complicated and burdensome. Moreover, learnable queries \cite{valanarasu2022transweather} or codebook priors \cite{ye2023adverse} are introduced to facilitate weather-specific feature learning. However, these methods neglect the shared characteristics across various weather conditions. To address this, Zhu \textit{et al.} \cite{zhu2023learning} propose a two-stage framework that separately extracts weather-general and weather-specific features. However, customized modifications of network architectures restrict its adaptability to unpredictable weather distortions in real-world scenarios. 

As the power of generative paradigms, the diffusion model succeeds in restoring realistic and natural images \cite{ho2020denoising, saharia2022image, ozdenizci2023restoring}. WeatherDiffusion \cite{ozdenizci2023restoring} is the first attempt to employ the diffusion model for adverse weather restoration, yet there remains room for further performance improvement. Firstly, sampling from pure Gaussian noise is unnecessary since the degraded image is available. Secondly, large sampling steps increase inference time. Finally, the potential correlations and diversities among distinct weather degradations are overlooked. Thus, \textbf{\emph{1. how to effectively and efficiently exploit weather-specific and shared features in the diffusion model for all-in-one adverse weather restoration is worth considering}}. Moreover, intricate and diverse weather conditions are often encountered in real-world scenarios, \textit{e.g.}, unseen or even hybrid weather scenarios. However, existing static networks struggle to generalize such complicated degradations. Hence, \textbf{\emph{2. how to design a flexible model that can adaptively generalize complex degradations requires further exploration.}} To address these challenges, we explore the degradation-aware adaptive diffusion priors for all-in-one weather restoration, termed DA$^{2}$Diff. It's an innovative all-in-one framework that harnesses the powerful perceptual capability of CLIP \cite{radford2021learning} to extract degradation-aware representations. These representations are dynamically integrated with various restoration experts, enabling the diffusion model to handle diverse weather degradations adaptively.

We build upon the diffusion paradigm \cite{zheng2024selective} with strong condition guidance and shared distribution term. It can overcome some limitations of WeatherDiffusion but overlooks the unique degradation characteristics of each weather type. The CLIP's powerful perception capability holds the potential for degradation-aware perception, motivating us to generate the degradation-aware priors by CLIP to the diffusion model. Unlike predefined text \cite{luo2023controlling} or prompt engineering \cite{jiang2025autodir} for describing degradation information, we design learnable prompts in CLIP to capture degradation-related features by aligning image-prompt pairs. Specifically, we separately employ pre-trained image and text encoders to encode the degraded images and learnable prompts into the CLIP latent space. By narrowing the disparity between the degraded images (\textit{i.e.}, rainy, hazy, or snowy images) and their corresponding weather-specific learnable prompt (\textit{i.e.}, rain, haze, or snow prompt) in the latent space, each learnable prompt contributes to the different weather degradation characteristics. The learned prompts are integrated into the diffusion model via the proposed weather-specific prompt guidance (WPG), enabling the model to effectively restore multiple weather types. To further boost the model's generalization to complex degradations in real-world scenarios, we develop a dynamic expert selection modulator (DESM) to adaptively assign relevant restoration experts to degraded images based on a dynamic weather-aware router, enhancing the model’s adaptability to diverse real-world degradations. Unlike \cite{shazeer2017outrageously} with a fixed number of activated experts, we dynamically adjust the number of activated experts for every input, improving computational efficiency and restoration performance. As exhibited in Fig. \ref{fig: Visual results}, our DA$^{2}$Diff can generalize to restore real-world weather degradations, yielding visually appealing results. Comprehensive experiments also demonstrate that DA$^{2}$Diff performs favorably against the state-of-the-art all-in-one weather removal approaches.

Overall, our main contributions are as follows:
\begin{itemize}
    \item We propose DA$^{2}$Diff, a novel diffusion paradigm that learns degradation-aware adaptive priors for all-in-one weather restoration, which is a new application of the large-scale vision-language model CLIP for learning weather-aware representations.
    \item We develop a degradation-aware prompt learning strategy that harnesses learnable prompts in CLIP to capture the distinctive characteristics among different weather degradations. The learned weather prompts are incorporated into the diffusion model via the designed weather-specific prompt guidance (WPG) module, making it possible to restore multiple weather degradations.
    \item We develop a dynamic expert selection modulator (DESM) that employs a dynamic weather-aware router to flexibly assign varying numbers of restoration experts for each degraded image, improving the model's adaptability to diverse degradations and computational efficiency.
\end{itemize}

The remainder of this paper is arranged as follows: Section \ref{section II} reviews the related work. Section \ref{section III} presents the preliminaries of the diffusion paradigm \cite{zheng2024selective}. In Section \ref{section IV}, we introduce the methodology of our DA$^{2}$Diff. Section \ref{section V} reports and analyzes the experimental results. Finally, the conclusion is summarized in Section \ref{section VI}. 

\section{Related Work}
\label{section II}
\subsection{Adverse Weather Removal}
\textbf{Single Weather Removal.} Due to distinct physical imaging principles among different weather conditions, previous works are dedicated to single weather restoration. \textit{For haze removal}, early efforts \cite{he2010single, cai2016dehazenet, li2017aod, zhang2018densely} employ hand-crafted priors or deep neural networks to estimate the parameter of physical model \cite{narasimhan2000chromatic}. Subsequently, learning-based methods directly restore haze-free images from hazy images using attention mechanisms \cite{liu2019griddehazenet}, GANs \cite{qu2019enhanced}, or Transformers \cite{song2023vision}. \textit{For rain removal}, a line of works focuses on rain streak removal with some techniques, such as recurrent network \cite{li2018recurrent}, spatial attention \cite{wang2019spatial}, or conditional VAEs \cite{du2020conditional}. The other line of work adopts attentive GANs \cite{qian2018attentive} or mathematical descriptions \cite{quan2019deep} to raindrop removal. \textit{For snow removal}, DesnowNet \cite{liu2018desnownet} is the first CNN-based method for image desnowing. JSTASR \cite{chen2020jstasr} develops a joint size and transparency-aware network to eliminate the veiling effect of snow. DDMSNet \cite{zhang2021deep} integrates semantic and geometric priors into a dense multi-scale network for better snow removal. Although these methods achieve excellent results in specific weather degradation, they suffer from noticeable performance deterioration when handling other weather conditions. 

\textbf{Multiple Weather Removal.} Several approaches \cite{zamir2021multi, tu2022maxim, zamir2022restormer, chen2022simple} explore general networks to tackle multiple degradations. For instance, MPRNet \cite{zamir2021multi} exploits a multi-stage strategy to refine restored images progressively. Restormer \cite{zamir2022restormer} introduces an efficient Transformer that captures global dependency features in channel dimension for effective image restoration. These general restoration networks support multiple weather removal within a single framework, whereas they need to train individual pre-trained weights for each weather type. 

Recent works \cite{li2020all, chen2022learning, valanarasu2022transweather, zhu2023learning, ozdenizci2023restoring, gao2023frequency, zhu2024mwformer, li2024multi} have developed a unified model for multiple weather restoration in an all-in-one manner. Among them, All-in-One \cite{li2020all} deploys multiple encoders to restore multiple weather conditions, each tailored for specific weather degradations. However, the high computational cost of multiple encoders and neural architecture search hinders its real-world applicability. TransWeather \cite{valanarasu2022transweather} incorporates learnable weather-type queries into Transformer decoder to learn weather-related degradation, yet it ignores the similar attributes among various weather degradations. Furthermore, WGWSNet \cite{zhu2023learning} adopts a two-stage training strategy to learn the general and specific characteristics of different weather degradations. However, it requires customized modifications of network architectures for specific tasks, limiting its architectural flexibility and generalization capabilities to unseen degradations. WeatherDiffusion \cite{ozdenizci2023restoring} is the first work that harnesses the diffusion model for adverse weather removal. However, weak condition guidance and slow inference speed hamper its effectiveness and efficiency in multi-weather restoration. Therefore, our research is dedicated to providing robust degradation-aware priors for the diffusion model, with fewer inference steps, enabling the model to restore diverse weather degradations adaptively.

\subsection{Vision-Language Model}
With the remarkable cross-modal representations and zero-shot capabilities, the large-scale vision-language model CLIP \cite{radford2021learning} is widely used in various tasks, such as image manipulation \cite{patashnik2021styleclip, wei2022hairclip}, image generation \cite{crowson2022vqgan}, dense prediction \cite{wang2022cris, rao2022denseclip}, and image restoration \cite{luo2023controlling, liang2023iterative, sun2024coser}. Taking image restoration tasks as an example, Luo \textit{et al.} \cite{luo2023controlling} propose DA-CLIP that controls CLIP to predict degradation types and generate clean content embeddings, aligned with the predefined text description. Yet, a simple text description of degradation types fails to convey the precise degradation information. Liang \textit{et al.} \cite{liang2023iterative} leverage CLIP priors for backlit image enhancement in an unsupervised manner, where positive and negative text prompts are designed to distinguish well-lit and backlit images. Sun \textit{et al.} \cite{sun2024coser} explore the potential of pre-trained CLIP image encoder to extract cognitive information of preprocessed low-resolution images for real-world image super-resolution. Unlike the above methods, we design a set of learnable prompts in CLIP to achieve different degradation representations, helping the diffusion model to perceive weather-specific characteristics.

\subsection{Sparse Mixture-of-Experts} 
Mixture of experts (MoE) assembles a series of sub-models with identical architecture (called experts) and performs conditional computation in an input-dependent manner \cite{sener2018multi, jacobs1991adaptive}. The sparse mixture of experts (SMoE) \cite{shazeer2017outrageously}, a variant of MoE, exploits a router mechanism to activate relevant experts selectively, improving the model's scalability and efficiency. SMoE is mainly employed in natural language processing \cite{du2022glam, shazeer2017outrageously} and computer vision \cite{enzweiler2011multilevel, riquelme2021scaling, ng2023botbuster, yang2024multi, zhang2024efficient}. The pioneering work \cite{riquelme2021scaling} of vision applications introduces the transformer-based SMoE for image recognition. Yang \textit{et al.} \cite{yang2024multi} propose a decoder-focused framework that introduces the generic convolution path and low-rank expert path to the
SMoE structure for multi-task dense prediction. Zhang \textit{et al.} \cite{zhang2024efficient} develop an efficient MoE architecture with two core components for adverse weather removal, \textit{i.e.}, uncertainty-aware router and feature modulated expert, significantly reducing computation overhead. In this work, we focus on dynamically adjusting the number of activated experts for every input based on a dynamic weather-aware routing mechanism, flexibly applying relevant experts to restore degraded images.

\section{Preliminaries}
\label{section III}
The diffusion paradigm \cite{zheng2024selective}, built upon standard T-step diffusion model \cite{ho2020denoising}, develops the selective hourglass mapping strategy equipped with strong condition guidance and shared distribution term. In the forward process, the transition distribution is formulated as follows: 
\begin{equation}
    q\left(I_{t}\mid I_{t-1}, I_{res}, I_{in}\right) = \mathcal{N} \left (I_{t}; I_{t-1} + \alpha_{t} I_{res} - \delta_{t} I_{in}, \beta_{t}^{2} \mathbf{I} \right) 
\end{equation}
where $I_{t}$ is the diffusive images at time step $t$, and $I_{res}$ refers to the residual between degraded image $I_{in}$ and clean image $I_{0}$, \textit{i.e.}, $I_{res}=I_{in}-I_{0}$. The $\mathcal{N} \left ( x;  \mu, \sigma  \right ) $ represents that data $x$ follows a normal distribution with mean $\mu$ and variance $\sigma$, and $\delta_{t} I_{in}$ is the shared distribution term. $\alpha_{t}$, $\beta_{t}$ and $\delta_{t}$ are noise coefficient of $I_{res}$, Gaussian noise, and shared distribution coefficient, respectively. Based on Markov chain and reparameterization technology \cite{kingma2013auto, kingma2019introduction}, the above equation is reformulated in closed form: 
\begin{equation}
    q\left(I_{t}\mid I_{0}, I_{res}, I_{in}\right) = \mathcal{N}\left (I_{t}; I_{0} + \bar{\alpha}_{t} I_{res} - \bar{\delta_{t}} I_{in}, \bar{\beta}_{t}^{2} \mathbf{I} \right)  
\end{equation}
\begin{equation}
\label{eq3}
    I_{t}= I_{0} + \bar{\alpha}_{t} I_{res} + \bar{\beta}_{t} \epsilon_{t} - \bar{\delta_{t}} I_{in} 
\end{equation}
where $\bar{\alpha}_{t} = \sum_{i=1}^{t} \alpha_{t}$, $\bar{\beta}_{t} = \sqrt{\sum_{i=1}^{t} \beta_{t}^2}$, $\bar{\delta}_{t} = \sum_{i=1}^{t} \delta_{t}$, and $\mathbf{\epsilon}_t \sim \mathcal{N}(\mathbf{0}, \mathbf{I})$. When $t \to T$, $\bar{\alpha}_{T}=1 $, $\bar{\delta}_{T} =0.9$, thereby formula \ref{eq3} could be rewritten as $I_{T} = \left ( 1-\bar{\delta}_{T} \right ) I_{in} + \bar{\beta}_{T}\epsilon_{T} = 0.1 I_{in} + \bar{\beta}_{T} \epsilon_{T}$. 

The reverse process is designed to reconstruct high-quality images from the noisy-carrying degraded images. Each iteration can be written as Markov Chain:
\begin{equation}
    p_{\theta}\left ( I_{t-1}\mid I_{t}, I_{in} \right ) = \mathcal{N}\left (I_{t-1}; u_{\theta}\left ( I_{t}, I_{in}, t \right ), \sigma_{t}^{2} \mathbf{I}\right )  
\end{equation}
where the mean $u_{\theta} \left ( I_{t}, I_{in}, t \right ) = I_{t} - \alpha_{t} I_{res}^\theta + \delta_{t} I_{in} - \frac{\beta_{t}^{2} }{\bar{\beta}_{t}} \epsilon_{t} ^\theta$ and variance $\sigma_{t} = \frac{\beta_{t} \bar{\beta}_{t-1}}{\bar{\beta}_{t}} $. The variable $I_{res} ^ \theta$ is predicted by the residual estimation network while variable $\epsilon_{t} ^ \theta$ is derived by $I_{res} ^ \theta$. By the implicit sampling strategy \cite{song2020denoising} and reparameterization technology, $I_{t-1}$ could be sampled from $I_{t}$ by: 
\begin{equation}
    I_{t-1} = I_{t} - \alpha_{t} I_{res}^\theta + \delta_{t} I_{in} 
\end{equation}
where the residual estimation value $I_{res} ^ \theta$ is optimized by following objective: 
\begin{equation}
    \mathcal{L}_{res}\left ( \theta  \right )  = \mathbb{E}_{t, I_{t}, I_{res}, I_{in}} \left [ \left \| I_{res} - I_{res}^{\theta}\left (I_{t}, I_{in}, t \right )   \right \| _{1} \right ] 
\end{equation}

\section{Methodology}
\label{section IV}
To better perceive degradation properties and adaptively tackle diverse degradations, we propose an innovative diffusion paradigm with degradation-aware adaptive priors for all-in-one weather restoration. DA$^{2}$Diff employs CLIP to extract degradation-aware features and dynamically integrates the features into various restoration experts, enabling the diffusion model to restore multiple weather degradations adaptively. Specifically, the disparate attributes of different degradations motivate us to utilize degradation-aware features for effective multi-weather restoration. Inspired by the powerful vision-language representation capabilities of CLIP, we apply it to learn a set of degradation-related prompts by imposing prompt-image similarity constraints in the CLIP space. These learned weather prompts are integrated into the diffusion model via the proposed weather-specific prompt guidance module, enabling the model to customize restoration schemes for each weather type. Furthermore, we design a dynamic expert selection modulator, which employs a dynamic weather-aware router to flexibly assign varying numbers of restoration experts for every degraded image, allowing the diffusion model to restore diverse degradations adaptively.

In this section, we first describe the overview of our method in Sec. \ref{overview}. Next, the first-stage degradation-aware prompt learning is introduced in Sec. \ref{stage1}. After that, the second-stage prompt guidance diffusion restoration is illustrated in Sec. \ref{stage2}. Finally, we detail the loss functions in Sec. \ref{loss function}.

\begin{figure*}[htbp] 
    \centering
    \includegraphics[width=1.0\linewidth]{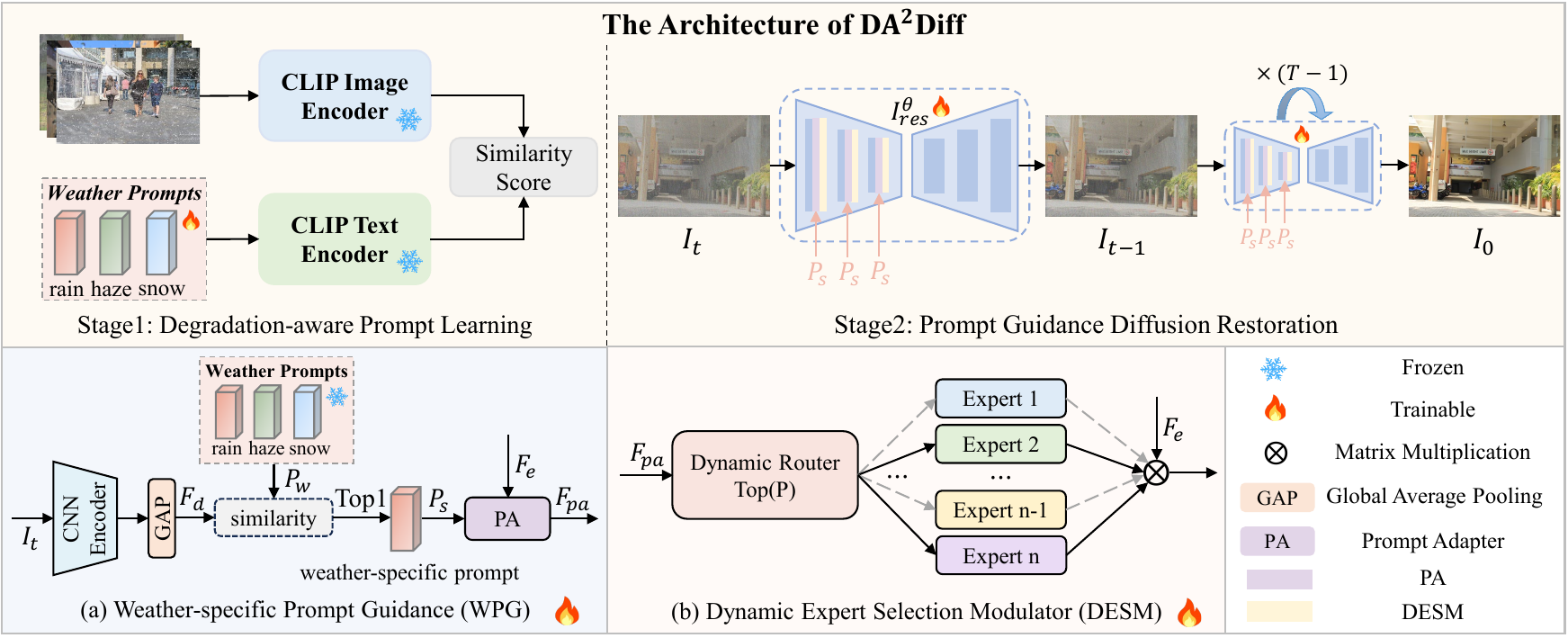}
	\caption{
    The overall architecture of DA$^{2}$Diff. It involves two stages: degradation-aware prompt learning and prompt guidance diffusion restoration. In the first stage, we freeze the parameters of the image encoder and text encoder in CLIP and learn the weather prompts through contrastive learning. In the second stage, the learned weather prompts $P_{w}$ provide degradation-aware adaptive priors for diffusion-based restoration by two core components: (a) WPG and (b) DESM. WPG selects the most similar prompt $P_{s}$ from $P_{w}$, which matches the $i$-th state images $I_{t}$. Then, the weather-specific prompt $P_{s}$ is integrated into each encoder layer of the residual estimation model by PA and DESM. PA embeds the prompt $P_{s}$ into the feature map $F_{e}$ to generate degradation-aware features $F_{pa}$. Based on $F_{pa}$, DESM dynamically dispatches relevant restoration experts for the feature map $F_{e}$. Note that $F_{e}$ represents the output features of each encoder layer in the residual estimation model.}
    \label{fig:overview}
\end{figure*}

\subsection{Overview}
\label{overview}
The overall architecture of DA$^{2}$Diff is illustrated in Fig. \ref{fig:overview}, which contains two stages: degradation-aware prompt learning and prompt guidance diffusion restoration. In the first stage, we leverage the vision-language model CLIP to learn a set of weather prompts. By narrowing the distance between the weather-specific degraded images (snowy, rainy, or hazy images) and their corresponding learnable prompt (snow, rain, or haze prompt) using contrastive loss, each prompt is tailored to capture a specific weather degradation. In the second stage, we propose two core components to provide degradation-aware adaptive priors for the diffusion model: weather-specific prompt guidance (WPG) and dynamic expert selection modulator (DESM). WPG selects the most similar prompt $P_{s}$, matched to the latent images $I_{t}$, from weather prompts. The degradation-aware prompt $P_{s}$ is then embedded into the output features $F_{e}$ of each encoder layer in the residual estimation model through the prompt adapter (PA). The detailed structure of PA is exhibited in Fig. \ref{fig:PA}. In DESM, based on the degradation-aware representations $F_{pa}$, the dynamic router computes a probability distribution over a set of experts and activates the relevant restoration experts. These activated experts then collaborate with the feature map $F_{e}$ to perform adaptive multi-weather restoration. 

\subsection{Degradation-aware Prompt Learning}
\label{stage1}
\textbf{Motivation.} As analyzed in \cite{zhu2023learning}, distinct weather degradations share common attributes, such as low contrast and color distortion. Meanwhile, they also exhibit unique characteristics, such as varying shapes and scales of atmospheric particles. Inspired by this wisdom, we explore how to extract both shared and weather-specific features within diffusion model for better all-in-one weather restoration. On the other hand, the novel diffusion paradigm \cite{zheng2024selective} achieves fewer sampling steps with strong condition guidance and extracts the shared features among different degradations with a shared distribution term. It motivates us to adopt the diffusion paradigm \cite{zheng2024selective} for all-in-one weather restoration. Although \cite{zheng2024selective} overcomes some limitations of the diffusion paradigm used in WeatherDiffusion \cite{ozdenizci2023restoring}, it ignores the unique degradation characteristics across different weather conditions. Recently, the large-scale vision-language model CLIP exhibits outstanding image-text representation capabilities, holding the potential for perceiving weather-specific characteristics. Therefore, we harness CLIP to construct a set of learnable weather prompts for extracting degradation-aware representations.

The process of degradation-aware prompt learning is illustrated in Fig. \ref{fig:overview}. In this stage, we freeze all parameters of the image encoder and text encoder in CLIP while training solely on the textual prompts. Our goal is to learn the degradation-aware text representations without updating the encoders. Concretely, we employ the pre-trained CLIP model to learn three types of weather prompts, aligned with three common weather types, \textit{i.e., snow, haze, and rain}. Given a snowy image $I_{s}$, a rainy image $I_{r}$, and a hazy image $I_{h}$, we randomly initialize snowy prompt $T_{s'}$, rainy prompt $T_{r'}$, and hazy prompt $T_{h'}$ with the form $\left [ X \right ]_{1} \left [ X \right ]_{2}\dots \left [ X \right ]_{N}$. All textual prompts $\in \mathbb{R}^{N\times512}$ where $N$ denotes the length of embedded tokens in each prompt. The snowy, rainy, and hazy images are passed through the fixed CLIP image encoder to extract the image features. Meanwhile, the textual prompts of snow, rain, and haze are fed into the fixed text encoder to obtain the text features. By adopting the cross entropy loss, the image features and text features are aligned in common CLIP latent space, allowing each learnable textual prompt to capture specific weather degradation. The cross entropy loss $\mathcal{L}_{ce}$ can be expressed as follows:
\begin{equation}
    \mathcal{L}_{ce}= -\frac{1}{3} \sum_{i\in\{s, r, h \}}\sum_{j\in\{s', r', h'\}}y_{ij} \log (\hat{y}_{ij})
\end{equation}
\begin{equation}
    \hat{y}_{ij}=\frac{e^{cos(\Phi_{image}(I_{i}), \Phi_{text}(T_{j}))}}{\sum_{k\in{\{s',r',h'\}}}e^{cos(\Phi_{image}(I_{i}),\Phi_{text}(T_{k}))}}
\end{equation}
where $y_{ij}$ is the label of the image $I_{i}$, here 0 is for rainy image $I_{r}$, 1 is for hazy image $I_{h}$, and 2 is for snowy image $I_{s}$. $\Phi_{image}(\cdot)$ and $\Phi_{text}(\cdot)$ represent the image encoder and text encoder of CLIP, respectively. 

\subsection{Prompt Guidance Diffusion Restoration}
\label{stage2}
\textbf{Motivation.} The complex and varied weather degradations in real-world scenarios pose challenges for restoration models. In this stage, we are dedicated to providing the degradation-aware adaptive priors for diffusion model, improving the model's adaptiveness to diverse weather conditions. SMoE \cite{shazeer2017outrageously} is a network architecture with a learnable gating mechanism, which sparsely routes input tokens to specialized expert sub-networks. This flexible design makes it possible to adaptively restore diverse degradations. Yet, the routing mechanism in SMoE activates the fixed number of relevant experts for every input, ignoring input complexity of distinct weather-distorted images. It is reasonable to allocate fewer experts for simple degradation inputs and more experts for complex degradation inputs. Therefore, we introduce a dynamic routing mechanism to dynamically adjust the number of activated experts for distinct degradation inputs. 

In this stage, we propose two core components to provide degradation-aware adaptive priors for diffusion model, \textit{i.e.}, WPG and DESM. WPG contributes to matching the most appropriate prompt $P_{s}$ for state images $I_{t}$ and enables the interaction between degradation-related prompt $P_{s}$ and the input features $F_{e}$. Note that we regard the output features of each encoder layer in the residual estimation model as $F_{e}$. DESM employs a dynamic weather-aware routing mechanism to adaptively assign a variable number of restoration experts, enabling the flexible restoration of relevant experts. Next, we describe the proposed WPG and DESM in detail.

\begin{figure}[t] 
    \centering
    \includegraphics[width=0.9\linewidth]{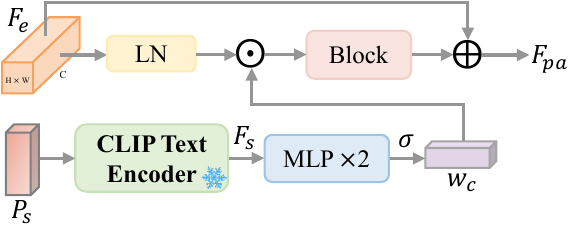}
	\caption{
        The architecture of prompt adapter (PA). PA aims to integrate the weather-specific prompt $P_{s}$ into the feature map $F_{e}$, providing degradation-aware guidance for diffusion model.}
	\label{fig:PA}
\end{figure}

\textbf{Weather-specific Prompt Guidance.} The detailed architecture of WPG is depicted in Fig. \ref{fig:overview} (a). Given a latent state $I_{t} \in \mathbb{R}^{H\times W \times C}$, WPG first extracts the shallow features $F_{d} \in \mathbb{R}^{1 \times D}$ by employing convolution operation and global average pooling, where $H \times W$ denotes the spatial resolution, while $C$ and $D$ represent the channel dimension of latent state $I_{t}$ and prompt embedding $P_{w}\in \mathbb{R}^{N \times D}$, respectively. Then, we calculate the cosine similarity between the shallow features $F_{d}$ and weather prompts $P_{w}$, and select the most relevant degradation prompt $P_{s}$ for input state $I_{t}$. Next, the selected degradation-related prompt $P_{s}$ is integrated at each encoder level of the residual estimation network via the prompt adapter, enabling the interaction with input features $F_{e}$. As illustrated in Fig. \ref{fig:PA}, PA feeds the degradation-related prompt $P_{s}$ into fixed CLIP text encoder $\mathcal{E}$ to obtain text embedding $F_{s}$. Then, $F_{s}$ applies a two-layer MLP and sigmoid operation $\sigma$ to generate a set of channel-wise weight vector $w_{c}$. We incorporate $w_{c}$ into input features $F_{e}$ along the channel dimension and employ the NAFBlock \cite{chen2022simple} to further enhance information encoding. The overall process of PA is formalized:
\begin{equation}
    F_{pa} = PA \left ( F_{e}, P_{s} \right ) = Block\left (  LN\left ( F_{e} \right )  \odot w_{c} \right ) + F_{e}   
\end{equation}
\begin{equation}
    w_{c} = Sigmoid \left ( MLP \left ( \mathcal{E} \left ( P_{s} \right )  \right )  \right )
\end{equation}
where LN is the layer normalization, $\odot$ represents element-wise multiplication, and $w_{c}$ refers to $c$-dimensional channel-wise weights based on the weather-specific prompt $P_{s}$. By integrating the degradation-aware prompt at each encoder level, the prompt adapter implicitly guides diffusion model to capture degradation-specific features.

\textbf{Dynamic Expert Selection Modulator.} To further enhance the model's adaptability to diverse weather conditions, we propose the DESM to dynamically combine restoration experts for various weather degradation restoration. As illustrated in Fig. \ref{fig:overview} (b), built upon the SMoE structure, we construct a set of restoration experts $\left \{ E_{1}, E_{2}, \dots, E_{n}\right \}$, where each expert is expertise at specific degradation type. By taking the degradation-related features $F_{pa}$ as input and evaluating the correlation between the degradation features $F_{pa}$ and specific experts $E_{i}$, the dynamic weather-aware router generates a set of weights for candidate experts and then sparsely activates the appropriate restoration experts via a dynamic routing mechanism Top(P). At last, the activated experts collaborate with the input features $F_{e}$ for adaptive restoration. Overall, the formulation of the DESM can be described as:
\begin{equation}
    DESM \left ( F_{e}, F_{pa} \right ) = \sum_{i=1}^{n} r_{i}\left ( F_{pa} \right ) e_{i}\left ( F_{e} \right ) 
\end{equation}
\begin{equation}
    r_{i}\left ( F_{pa} \right ) = Top \left ( P \right )  \left ( Softmax\left ( F_{pa}^{'} \right )  \right )
\end{equation}
\begin{equation}
\label{Eq13}
    F_{pa}^{'} = F_{pa} W_{g} + \mathcal{N} \left ( 0, 1 \right ) Softplus \left ( F_{pa} W_{noise} \right ) 
\end{equation}
where $F_{pa}$ represents the feature output of WPG, $r_{i}\left ( \cdot \right )$ is the router weight of $i$-th expert, and $e_{i}\left ( \cdot \right )$ denotes the features extracted by $i$-th expert network. Here, we adopt the feed forward networks as expert networks. Unlike the fixed Top-K routing in SMoE, $Top \left ( P \right )$ is a dynamic routing mechanism that activates a variable number of experts with higher routing scores until their cumulative scores surpass the threshold $P \in \left [ 0, 1 \right ]$. The larger values of $P$ indicate the more experts are activated. Note that the weights of experts without activation are set to 0. To improve the diversity and robustness of expert selection, a tunable Gaussian noise is added in the Softmax function (see Eq. \ref{Eq13}). The $W_{g}$ and $W_{noise}$ represent the learnable weight matrix of the input signal and random noise, respectively. $Softplus \left ( \cdot  \right )$ is the smooth approximation of ReLU function. The dynamic weather-aware routing mechanism dynamically assigns relevant experts to each input feature, making it possible to restore diverse degradations adaptively. 

\subsection{Loss Function}
\label{loss function}
In the first stage, we adopt cross entropy loss $\mathcal{L}_{ce}$ to learn a set of degradation-related textual prompts. In the second stage, apart from the residual estimation loss $\mathcal{L}_{res}$, we also employ the load balance loss $\mathcal{L}_{balance}$ \cite{fedus2022switch}. The gating function in SMoE structure often leads to a load imbalance problem \cite{shazeer2017outrageously}, where few same experts are repeatedly activated while others remain underutilized. To encourage different experts to process roughly equal numbers of samples, we apply the loss $\mathcal{L}_{balance}$ to ensure a balanced load among the experts. Given n experts and a batch $\mathcal{B}$ with $S$ samples, the load balance loss can be formalized as follows:
\begin{equation}
    \mathcal{L}_{balance} = n \sum_{i=1}^{n} f_{i} \cdot P_{i}
\end{equation}
\begin{equation}
    f_{i} = \frac{1}{S} \sum_{x \in \mathcal{B}} \mathbbm{1} \left \{ argmax \left ( p\left ( x \right ) = i \right ) \right \} 
\end{equation}
\begin{equation}
    P_{i} = \frac{1}{S} \sum_{x\in \mathcal{B}} p_{i}\left ( x \right ) 
\end{equation}
where $f_{i}$ is the fraction of samples assigned to the $i$-th expert. $P_{i}$ is the fraction of router probability dispatched to the $i$-th expert. $\mathbbm{1} \left \{ argmax \left ( p\left ( x \right ) = i  \right ) \right \} $ defines an indicator function. It returns 1 when condition $argmax \left ( p\left ( x \right ) = i \right )$ is satisfied; otherwise, it returns 0. $p_{i} \left ( x \right ) $ is the router probability for the $i$-th expert given sample $x$. 

Hence, the total loss of the prompt guidance diffusion restoration stage is defined as:
\begin{equation}
    \mathcal{L}_{total} = \mathcal{L}_{res} + \lambda \mathcal{L}_{balance} 
\end{equation}
where $\lambda$ is a hyperparameter to balance $\mathcal{L}_{res}$ and $\mathcal{L}_{balance}$. In our experiments, $\lambda$ is set to 0.01.

\section{Experiments}
\label{section V}
In this section, we first present the benchmark datasets, implementation details, and evaluation settings in our experiments. Next, we perform the performance comparisons against the state-of-the-art approaches on both synthetic and real-world datasets. Finally, we conduct ablation studies to analyze the effectiveness of different designs in our model.
\subsection{Experimental Setup}
\textbf{Datasets.} For fair comparisons, we evaluate the performance of DA$^2$Diff on the All-weather \cite{valanarasu2022transweather} benchmark, as the previous methods \cite{li2020all, valanarasu2022transweather, ozdenizci2023restoring}. All-weather is a combination of three subsets derived from the datasets: Raindrop \cite{qian2018attentive}, Outdoor-Rain \cite{li2019heavy}, and Snow100K \cite{liu2018desnownet}, with images collected under raindrop, heavy rain with rain streaks and haze, and snowy conditions. The training set contains 18,069 image pairs, including 818 from Raindrop \cite{qian2018attentive}, 8,250 from Outdoor-Rain \cite{li2019heavy}, and 9,001 from Snow100K \cite{liu2018desnownet}. The test set encompasses 58 images from RainDrop test set \cite{qian2018attentive}, 750 images from Test1 \cite{li2019heavy}, and 16,081 images from Snow100K-L test set \cite{liu2018desnownet}. To further assess the model's generalization capabilities in real-world scenarios, we train models on the synthetic benchmark \cite{valanarasu2022transweather} and evaluate them on two real datasets: Snow100K-real \cite{liu2018desnownet} and RainDS-real \cite{quan2021removing}. Snow100K-real comprises 1,329 real-world snowy images. RainDS-real contains 450 training images and 294 testing images, captured under real raindrop and rain streak conditions. 

\textbf{Implementation Details.} All experiments are conducted on one RTX 3090 GPU with PyTorch \cite{paszke2019pytorch} framework. In the first stage, we utilize ViT-B/32 as the pre-trained CLIP image encoder and set the length of embedded tokens in each learnable prompt to 16. We train the weather prompts using the Adam optimizer over 8k iterations, with a learning rate of 5e$^{-6}$ and a batch size of 64. The input images are resized to 224 × 224 and augmentation techniques like random flipping, zooming, and rotation are adopted. In the second stage, we adopt the diffusion paradigm DiffUIR-L as the backbone. We train our model for 400k iterations using the Adam optimizer ($\beta_{1} = 0.9, \beta_{2} = 0.99$), with a batch size of 6 and a learning rate of 8e$^{-5}$. The images are randomly cropped to 256 × 256 and adopted random flipping for data augmentation. We configure the total number of experts $n$ to 4 and set the threshold $P$ to 0.4. To enhance the training stability of the diffusion model, the exponential moving average (EMA) strategy \cite{song2020improved} weighted at 0.995 is employed. Additionally, we use the implicit sampling strategy \cite{song2020denoising} to accelerate the sampling process and set the sampling steps to 3. 

\textbf{Evaluation Settings.} We adopt the widely used metrics PSNR \cite{assessment2004error} and SSIM \cite{assessment2004error} to assess the restored image quality. PSNR measures the pixel-wise error between the restored images and ground-truth images, while SSIM assesses image similarity of luminance, contrast, and structure. Higher scores of PSNR and SSIM commonly indicate superior performance. Additionally, we utilize the three no-referenced metrics for perceptual quality evaluation without reference images, \textit{i.e.}, NIQE \cite{zhang2015feature}, CLIP-IQA \cite{wang2023exploring}, and MANIQA \cite{yang2022maniqa}. Lower scores of NIQE mean better results, while higher scores of CLIP-IQA and MANIQA represent more promising results.

\begin{table}[htbp]
        \centering
        \caption{Quantitative comparisons against state-of-the-art methods on the Raindrop \cite{qian2018attentive} test set. The upper halves of the table present the weather-specific and general restoration results, while the lower halves report the comparison results with recent all-in-one weather removal methods. \textbf{Bold} and \underline{underlined} indicate the $1^{st}$ and $2^{nd}$ ranks, respectively.}
        \label{tab:raindrop}
        \begin{threeparttable}
		\footnotesize
		\centering
		\setlength{\tabcolsep}{1.0mm}{
			\begin{tabular}{cccccc}
				\toprule
				    \multirow{2}{*}{Method} &
                        \multirow{2}{*}{Publication} &
				    \multicolumn{2}{c}{RainDrop \cite{qian2018attentive}} \cr
				    \cmidrule(lr){5-6} & 
                        & PSNR$\uparrow$ & SSIM$\uparrow$ \cr
				\midrule
                        pix2pix \cite{isola2017image} & CVPR'17 & 28.02 & 0.8547 \\
                        DuRN \cite{liu2019dual}  & CVPR'19 & 31.24 & 0.9259 \\
                        AttentiveGAN \cite{qian2018attentive} & CVPR'18 & 31.59 & 0.9170 \\
                        RaindropAttn \cite{quan2019deep}& ICCV'19 & 31.44 & 0.9263 \\
                        CCN \cite{quan2021removing} & CVPR'21 & 31.34 & 0.9286 \\
                        IDT \cite{xiao2022image} & TPAMI'22 & 31.87 & 0.9313 \\
                        MAXIM \cite{tu2022maxim} & CVPR'22 & 31.87 & 0.9352 \\
                        Restormer \cite{zamir2022restormer}& CVPR'22 & 32.18 & 0.9408 \\
                        UDR-S$^{2}$Former \cite{chen2023sparse} & ICCV'23 & \underline{32.64} & \underline{0.9427} \\
                    \midrule
                        All-in-One \cite{li2020all} & CVPR'20 & 31.12 & 0.9268 \\
                        TransWeather \cite{valanarasu2022transweather} & CVPR'22 & 30.17 & 0.9157 \\
                        TUM \cite{chen2022learning}& CVPR'22 & 31.81 & 0.9309 \\
                        WGWSNet \cite{zhu2023learning} & CVPR'23 & 32.38 & 0.9378 \\
                        WeatherDiff\(_{64}\) \cite{ozdenizci2023restoring} & TPAMI'23 & 30.71 & 0.9312 \\
                        WeatherDiff\(_{128}\) \cite{ozdenizci2023restoring} & TPAMI'23 & 29.66 & 0.9225 \\
                        DiffUIR-L \cite{zheng2024selective} & CVPR'24 & 31.90 & 0.9368 \\
                        MW-ConvNet \cite{li2024multi} & TCSVT'24 & 31.18 & 0.9399 \\
                        MWFormer \cite{zhu2024mwformer} & TIP'24 & 31.73 & 0.9254 \\
                        Ours & - & \textbf{33.01} & \textbf{0.9451} \\
				\bottomrule
			\end{tabular}
		}
	\end{threeparttable}
\end{table}

\begin{table}[htbp]
        \centering
        \caption{Quantitative comparisons with state-of-the-art methods on the Test1 (Rain + Haze) \cite{li2019heavy} dataset.}
        \label{tab:outdoor-rain}
        \begin{threeparttable}
		\footnotesize
		\centering
		\setlength{\tabcolsep}{1.0mm}{
			\begin{tabular}{cccccc}
				\toprule
				    \multirow{2}{*}{Method}&
                        \multirow{2}{*}{Publication} &
				    \multicolumn{2}{c}{Outdoor-Rain \cite{li2019heavy}} \cr
				    \cmidrule(lr){5-6}&
			             & PSNR$\uparrow$ & SSIM$\uparrow$ \cr
				    \midrule
                        CycleGAN \cite{zhu2017unpaired} & ICCV'17 & 17.62 & 0.6560 \\
                        pix2pix \cite{isola2017image}& CVPR'17 & 19.09 & 0.7100 \\
                        HRGAN \cite{li2019heavy}& CVPR'19 & 21.56 & 0.8550 \\
                        PCNet \cite{jiang2021rain}& TIP'21 & 26.19 & 0.9015 \\
                        MPRNet \cite{zamir2021multi}& CVPR'21 & 28.03 & 0.9192 \\
                        NAFNet \cite{chen2022simple}& ECCV'22 & 29.59 & 0.9027 \\
                        Restormer \cite{zamir2022restormer} & CVPR'22 & 30.03 & 0.9215 \\
                        \midrule
                        All-in-One \cite{li2020all} & CVPR'20 & 24.71 & 0.8980 \\
                        TransWeather \cite{valanarasu2022transweather} & CVPR'22 & 28.83 & 0.9000 \\
                        TUM \cite{chen2022learning} & CVPR'22 & 29.27  & 0.9147  \\
                        WGWSNet \cite{zhu2023learning} & CVPR'23 & 29.32 & 0.9207 \\
                        WeatherDiff\(_{64}\) \cite{ozdenizci2023restoring} & TPAMI'23 & 29.64 & 0.9312 \\
                        WeatherDiff\(_{128}\) \cite{ozdenizci2023restoring} & TPAMI'23 & 29.72 & 0.9216 \\
                        DiffUIR-L \cite{zheng2024selective} & CVPR'24 & \underline{30.89} & 0.9231 \\
                        MW-ConvNet \cite{li2024multi} & TCSVT'24 & 30.78 & \textbf{0.9489} \\
                        MWFormer \cite{zhu2024mwformer} & TIP'24 & 30.24 & 0.9111 \\
                        Ours & - & \textbf{31.58} & \underline{0.9388} \\
				\bottomrule
			\end{tabular}
		}
	\end{threeparttable}
\end{table}

\begin{table}[htbp]
        \centering
        \caption{Quantitative comparisons with state-of-the-art methods on the Snow100K-L \cite{liu2018desnownet} test set.}
        \label{tab:snow}
        \begin{threeparttable}
		\footnotesize
		\centering
		\setlength{\tabcolsep}{1.0mm}{
			\begin{tabular}{cccccc}
				\toprule
				    \multirow{2}{*}{Method}&
                        \multirow{2}{*}{Publication} &
				    \multicolumn{2}{c}{Snow100K-L \cite{liu2018desnownet}} \cr
				    \cmidrule(lr){5-6}&
			             & PSNR$\uparrow$ & SSIM$\uparrow$ \cr
				\midrule
                        SPANet \cite{wang2019spatial} & CVPR'19 & 23.70 & 0.7930 \\
                        RESCAN \cite{li2018recurrent} & ECCV'18 & 26.08 & 0.8108 \\
                        DesnowNet \cite{liu2018desnownet} & TIP'18 &  27.17 & 0.8983 \\
                        JSTASR \cite{chen2020jstasr} & ECCV'20 & 25.32 & 0.8076 \\
                        DDMSNet \cite{zhang2021deep} & TIP'21 & 28.85 & 0.8772 \\
                        MPRNet \cite{zamir2021multi} & CVPR'21 & 29.76 & 0.8949 \\
                        NAFNet \cite{chen2022simple} & ECCV'22 & 30.06 & 0.9017 \\
                        Restormer \cite{zamir2022restormer} & CVPR'22 & 30.52 & 0.9092 \\
                    \midrule
                        All-in-One \cite{li2020all} & CVPR'20 & 28.33 & 0.8820 \\
                        TransWeather \cite{valanarasu2022transweather} & CVPR'22 &  29.31 & 0.8879 \\
                        TUM \cite{chen2022learning} & CVPR'22 & 30.24 & 0.9020 \\
                        WGWSNet \cite{zhu2023learning} & CVPR'23 & 30.16 & 0.9007 \\
                        WeatherDiff\(_{64}\) \cite{ozdenizci2023restoring} & TPAMI'23 &  30.09 & 0.9041 \\
                        WeatherDiff\(_{128}\) \cite{ozdenizci2023restoring} & TPAMI'23 &  29.58 & 0.8941 \\
                        DiffUIR-L \cite{zheng2024selective} & CVPR'24 & 30.64 & 0.9082 \\
                        MW-ConvNet \cite{li2024multi} & TCSVT'24 & \underline{30.92} & \textbf{0.9227} \\
                        MWFormer \cite{zhu2024mwformer} & TIP'24 & 30.70 & 0.9060 \\
                        Ours & - & \textbf{31.42} & \underline{0.9158} \\
				\bottomrule
			\end{tabular}
		}
	\end{threeparttable}
\end{table}

\subsection{Comparison with State-of-the-art Methods}
\textbf {Results on Synthetic Dataset.} We evaluate our DA$^{2}$Diff against various weather removal approaches, including \textit{weather-specific methods, general methods, and all-in-one weather restoration methods.} Specifically, for raindrop removal, the comparison includes pix2pix \cite{isola2017image}, DuRN \cite{liu2019dual}, AttentiveGAN \cite{qian2018attentive}, RaindropAttn \cite{quan2019deep}, CCN \cite{quan2021removing}, IDT \cite{xiao2022image}, and UDR-S$^{2}$Former \cite{chen2023sparse}. For rain + haze removal, we evaluate against the methods CycleGAN \cite{zhu2017unpaired}, pix2pix \cite{isola2017image}, HRGAN \cite{li2019heavy}, and PCNet \cite{jiang2021rain}. For snow removal, we compare with SPANet \cite{wang2019spatial}, RESCAN \cite{li2018recurrent}, DesnowNet \cite{liu2018desnownet}, JSTASR \cite{chen2020jstasr}, and DDMSNet \cite{zhang2021deep}. Moreover, we compare our method with general weather restoration methods, including MAXIM \cite{tu2022maxim}, Restormer \cite{zamir2022restormer}, MPRNet \cite{zamir2021multi}, and NAFNet \cite{chen2022simple}, which employ a single model to tackle multiple weather degradations with task-specific pre-trained weight. Furthermore, we perform the comparisons with all-in-one weather restoration approaches: All-in-One \cite{li2020all}, TransWeather \cite{valanarasu2022transweather}, TUM \cite{chen2022learning}, WGWSNet \cite{zhu2023learning}, WeatherDiff \cite{ozdenizci2023restoring}, MW-ConvNet \cite{li2024multi}, and MWFormer \cite{zhu2024mwformer}. Similar to \cite{li2020all, valanarasu2022transweather}, our method is trained on the mixed dataset \cite{valanarasu2022transweather} and tested on the specific dataset. 

Table \ref{tab:raindrop}, \ref{tab:outdoor-rain}, and \ref{tab:snow} report the quantitative comparison results for raindrop removal, deraining and dehazing, and image desnowing respectively. As reported, our method outperforms both weather-specific methods and general methods. This can be attributed to the successful application of degradation-aware prompts in CLIP, enabling the model to perceive different weather degradations for better multi-weather restoration. Compared to recent all-in-one approaches, DA$^{2}$Diff achieves superior results on the RainDrop test set by a significant margin (please refer to the table \ref{tab:raindrop}). It is particularly noteworthy that DA$^{2}$Diff exhibits noticeable improvement in PSNR/SSIM over the baseline diffusion paradigm DiffUIR-L, with gains of 1.11dB/0.0083. In addition, DA$^{2}$Diff achieves the highest PSNR on other test sets (please refer to the table \ref{tab:outdoor-rain} and \ref{tab:snow}), indicating superior restoration fidelity. In terms of SSIM, DA$^{2}$Diff obtains impressive results and ranks second among the eight all-in-one weather removal methods. 

\begin{figure*}[htbp] \centering
    \includegraphics[width=0.98\linewidth]{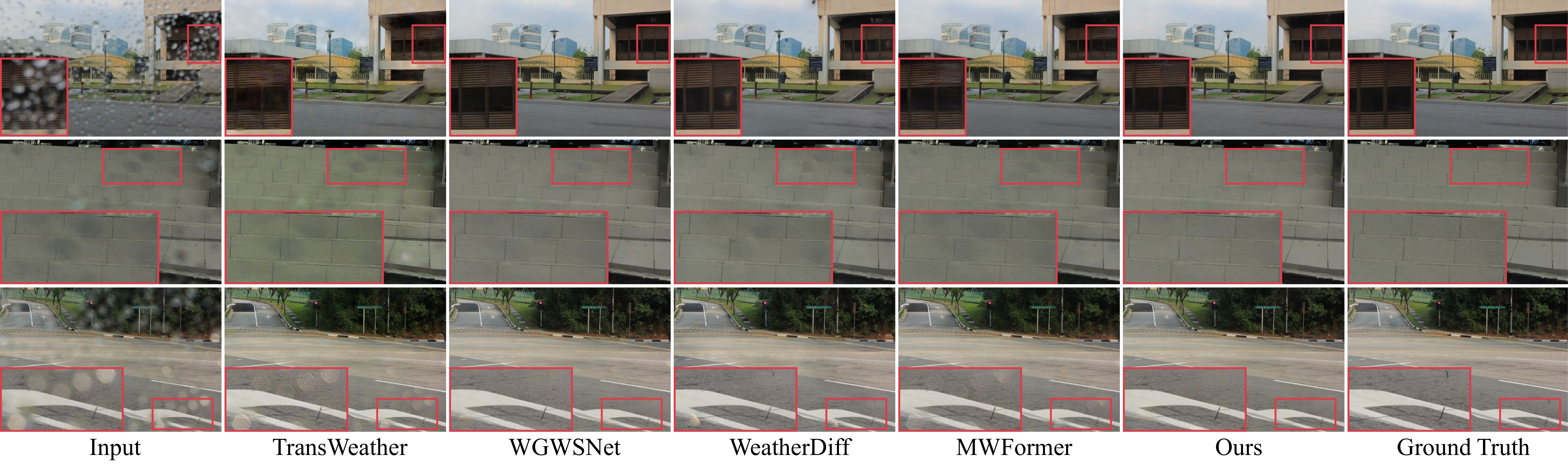}
 	\caption{Visual comparisons on the RainDrop \cite{qian2018attentive} test set. The region within the red box is zoomed for better comparison.}
   \label{fig:raindrop}
\end{figure*}

\begin{figure*}[htbp] \centering
 	\includegraphics[width=0.98\linewidth]{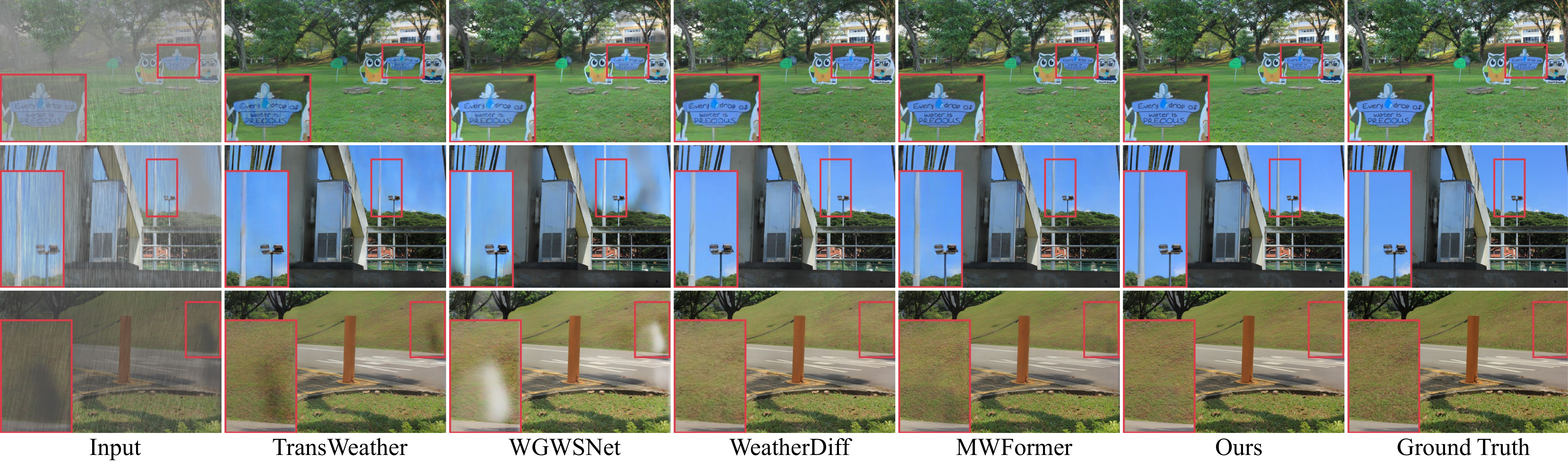}
 	\caption{Visual comparisons on the Test1 \cite{li2019heavy} (rain + haze) set. The region within the red box is zoomed for better comparison.}
   \label{fig:test1}
\end{figure*}

\begin{figure*}[htbp] \centering
 	\includegraphics[width=0.98\linewidth]{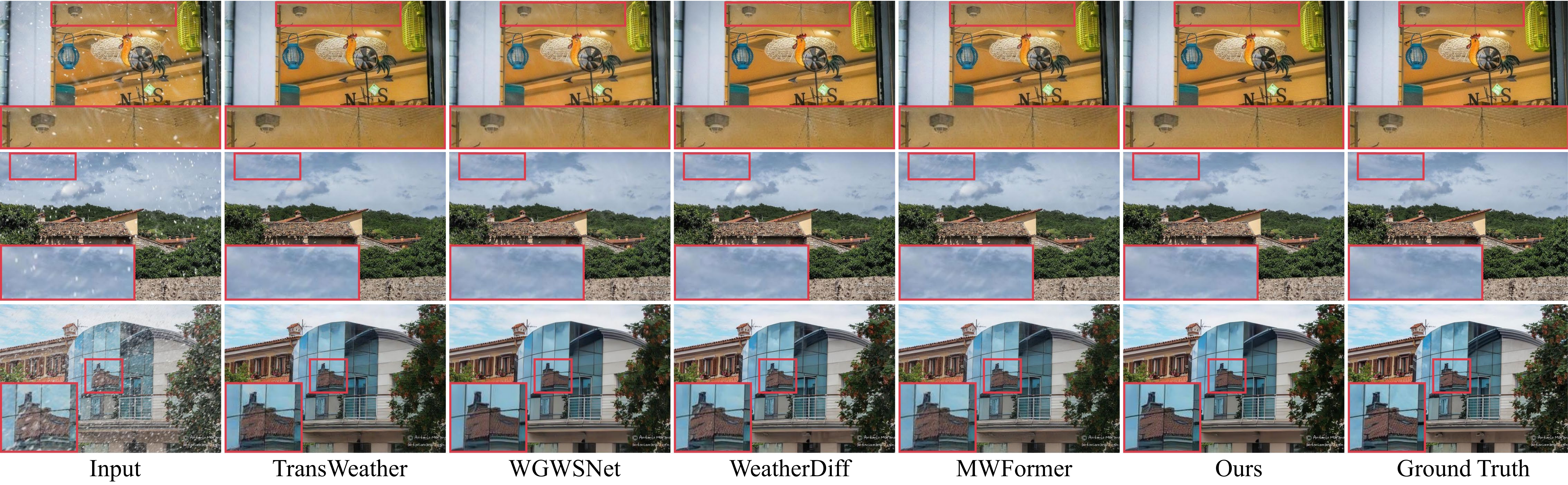}
 	\caption{Visual comparisons on the Snow100K-L \cite{liu2018desnownet} test set. The region within the red box is zoomed for better comparison.}
   \label{fig:snow100k-l}
\end{figure*}

Figs. \ref{fig:raindrop}-\ref{fig:snow100k-l} showcase the visual comparisons on each benchmark dataset respectively. For raindrop removal shown in Fig. \ref{fig:raindrop}, TransWeather and MWFormer cannot completely remove the raindrops (see the second and third row) and fail to recover the details (see the first row). WeatherDiff produces unexpected white artifacts (see the window in the first row). For image dehazing and deraining shown in Fig. \ref{fig:test1}, TransWeather and WGWSNet fail to restore the regions affected by dense black haze. MWFormer suffers from haze residuals (see the third row) and distorts image details (see the telephone pole in the second row). For image desnowing shown in Fig. \ref{fig:snow100k-l}, all compared methods retain some snow and exhibit limited ability to recover the texture details. In contrast, our method removes these weather degradations more thoroughly and preserves finer details, yielding visually pleasing results.

\begin{figure*}[htbp] \centering
 	\includegraphics[width=0.98\linewidth]{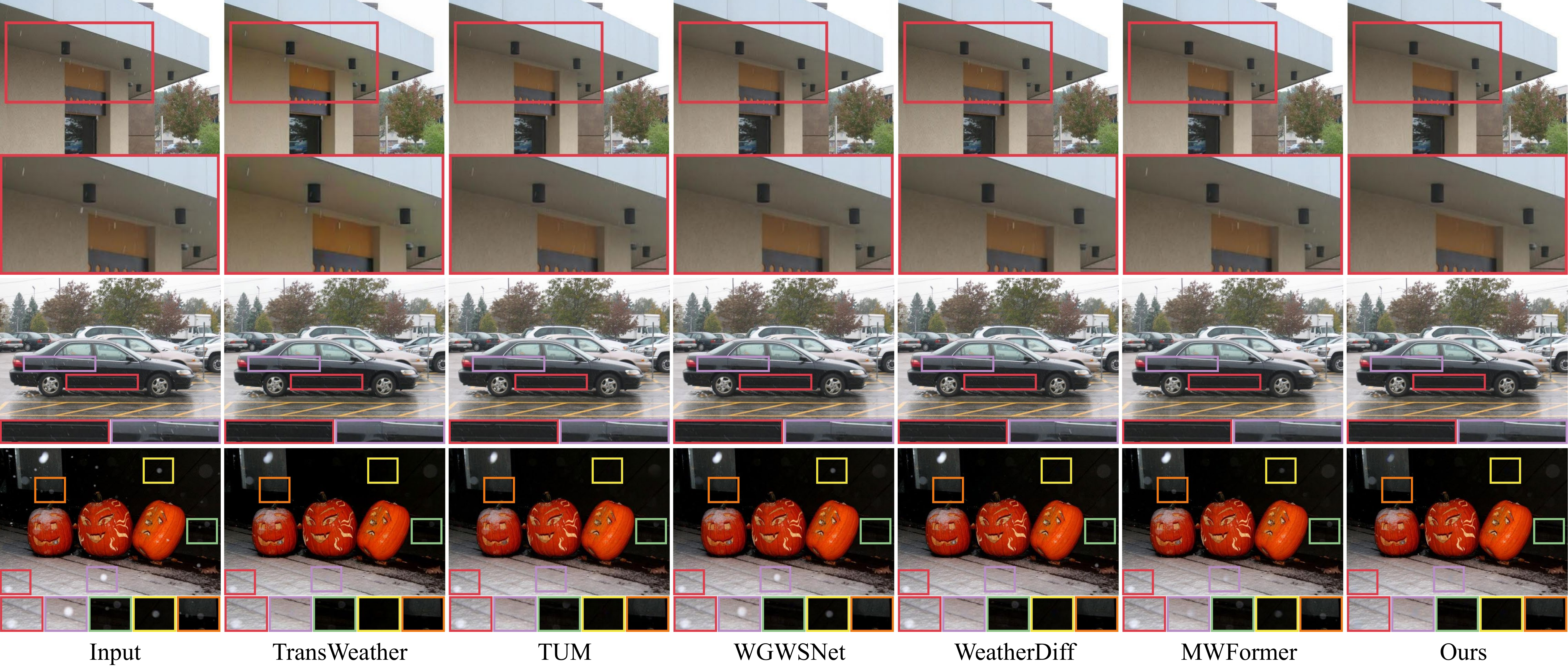}
 	\caption{Visual comparisons on the Snow100K-real \cite{liu2018desnownet} dataset. Our method can remove snowflakes successfully and generate more natural images.}
   \label{fig:snow100k-real}
\end{figure*}

\begin{figure*}[htbp] \centering
 	\includegraphics[width=0.98\linewidth]{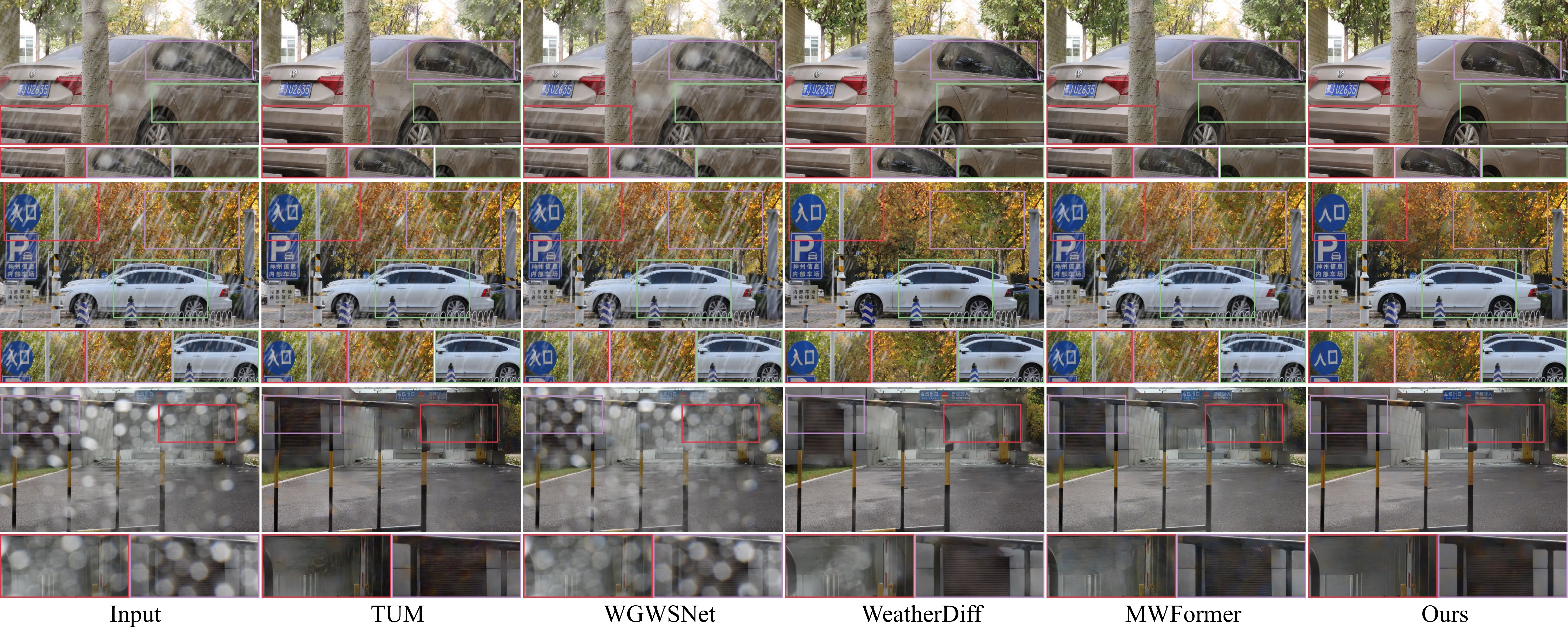}
 	\caption{Visual comparisons on the RainDS-real \cite{quan2021removing} test set. Our method produces visually pleasing results with fewer degradation residuals and finer details.}
   \label{fig:rainds}
\end{figure*}

\textbf{Results on Real-World Dataset.} To further assess the generalization ability of DA$^{2}$Diff on real-world images, we train our model on the synthetic dataset \cite{valanarasu2022transweather} and evaluate it on two unseen real-world datasets: Snow100K-real and RainDS-real. For fair comparisons, the other compared methods are also trained on the same synthetic dataset \cite{valanarasu2022transweather} and evaluated on the two test sets. Table \ref{Tab:snow100k-real} reports the averaged NIQE, CLIP-IQA, and MANIQA values of different algorithms on Snow100K-real dataset. As shown, our DA$^{2}$Diff obtains the highest scores in CLIP-IQA and MANIQA metrics and achieves the second-best performance in NIQE metric, indicating the superior adaptiveness of our method on real-world weather restoration.

\begin{table}[htbp]
	\centering
	\footnotesize
	\caption{Quantitative comparisons (niqe/clip-iqa/maniqa) on the Snow100K-real \cite{liu2018desnownet} dataset}
	\label{Tab:snow100k-real}
	\begin{threeparttable}
		\centering
		\setlength{\tabcolsep}{1.0mm}{
			\begin{tabular}{cccccc}
				\toprule
				\multirow{1}{*}{Method}&
				\multirow{1}{*}{NIQE $\downarrow$}&
				\multirow{1}{*}{CLIP-IQA $\uparrow$}&
				\multirow{1}{*}{MANIQA $\uparrow$} \cr
				\midrule
				TransWeather \cite{valanarasu2022transweather} & 3.0831 & 0.5057  & 0.3904\cr
				TUM \cite{chen2022learning} & 3.0659 & 0.4730 & 0.3778 \cr
                    WGWSNet \cite{zhu2023learning} & 3.0460 & 0.4812 & 0.3854 \cr
                    WeatherDiff \cite{ozdenizci2023restoring} & 3.0032 & \underline{0.5078} & \underline{0.3990} \cr
                    MWFormer \cite{zhu2024mwformer} & \textbf{2.9510} & 0.4937 & 0.3861\cr
                    \midrule
                    Ours & \underline{2.9586} & \textbf{0.5151} & \textbf{0.3996}\cr
				\bottomrule
			\end{tabular}
		}
	\end{threeparttable}
\end{table}

Fig. \ref{fig:snow100k-real} presents the visual comparisons of different all-in-one weather restoration methods on the Snow100K-real \cite{liu2018desnownet} dataset. It can be observed that our method removes the snowflakes with diverse sizes and shapes more thoroughly (particularly in the third row), while other methods exhibit snowflake residuals to some extent. Taking the third row as an example, we observe that all compared methods fail to remove the snowflake in the lower-left corner of the image. This may be because other methods interpret this snowflake as scene content and wrongly preserve it. Moreover, some flare artifacts caused by snow particles also are restored effectively by our method. These observations underscore the robustness and adaptiveness of DA$^{2}$Diff to generalize training-unseen real-world snow degradations.

Fig. \ref{fig:rainds} illustrates the visual comparison results on the RainDS-real \cite{quan2021removing} test set, which contains the hybrid degradations of raindrops and rain streaks. As a display, our method exhibits noticeable superiority over other methods, even in hybrid weather-induced degradations. In detail, our method almost removes the all raindrops and rain streaks from the input image in the first row, thanks to the successful application of the dynamic mixture-of-experts structure. Even in heavy rain conditions with low visibility (see the second and third row), our method still removes most of the raindrops or rain streaks, while restoring fine details. The visual results demonstrate that our DA$^{2}$Diff can generalize well to the complex weather degradations in real-world scenarios.

\subsection{Ablation Study}
\textbf{Effectiveness of Different Components.} To validate the efficacy of DA$^{2}$Diff, we conduct comprehensive ablation studies about its key components: the weather-specific prompt guidance (WPG), dynamic expert selection modulator (DESM), and load balance loss $\mathcal{L}_{balance}$. Here, we adopt the diffusion paradigm DiffUIR-L as the baseline. By progressively adding the components into the baseline model, several variants are constructed as follows: 
\begin{enumerate}
\item  baseline + WPG $\rightarrow$ $V_1$, 
\item  $V_1$ + DESM  $\rightarrow$ $V_2$,
\item  $V_2$ + $\mathcal{L}_{balance}$ $\rightarrow$ $V_3$ (full model).
\end{enumerate}
All these variants are trained in the same configurations as previously described and tested on RainDrop test set. The evaluation results of these variants are presented in Table \ref{Tab:ablation1}.

\begin{table}[htbp]
	\centering
	\footnotesize
	\caption{Ablation study of different components on RainDrop dataset.}
	\label{Tab:ablation1}
	\begin{threeparttable}
		\centering
		\setlength{\tabcolsep}{1.0mm}{
			\begin{tabular}{ccccc}
				\toprule
		    	\multirow{1}{*}{Variants}&
				\multirow{1}{*}{Baseline}&
				\multirow{1}{*}{$V_1$}&
				\multirow{1}{*}{$V_2$}&
				\multirow{1}{*}{$V_3$}\cr
				\midrule
				    Weather-specific Prompt Guidance & w/o & \checkmark & \checkmark & \checkmark  \cr
				    Dynamic Expert Selection Modulator & w/o & w/o  & \checkmark & \checkmark  \cr
				    Load Balance Loss & w/o & w/o & w/o & \checkmark \cr
				\midrule
				PSNR $\uparrow$ & 31.90 & 32.72 & 32.97     &  \textbf{33.01} \cr
				SSIM $\uparrow$ & 0.9368 & 0.9418 &         0.9429 & \textbf{0.9451} \cr
				\bottomrule
			\end{tabular}
		}
	\end{threeparttable}
\end{table} 

As shown, each component contributes to improving the restoration performance. Specifically, incorporating the WPG achieves advanced performance over the baseline model, with gains of 0.82dB in PSNR and 0.005 in SSIM, verifying the effectiveness of CLIP for perceiving degradation-related information. The introduction of DESM improves the performance in terms of PSNR and SSIM, owing to the sparse mixture-of-experts architecture with a dynamic degradation-aware routing mechanism. The application of load balance loss also contributes to performance gains, highlighting the importance of balanced expert loading. If we apply all components, the results will outperform other variants, confirming the effectiveness and necessity of each component.   

\textbf{Effectiveness of WPG.} The WPG is designed to select the most similar prompt aligned with the input state from a set of learnable weather prompts. After that, the selected prompt is embedded into each encoder layer of the residual estimation model via the prompt adapter. To investigate the effect of different designs in WPG on restoration performance, we conduct ablation experiments from three aspects: prompt configurations, integration strategy, and embedding positions. As presented in Table \ref{Tab:ablation2}, when we replace the learnable weather prompts with predefined text prompts, such as ``This is a rainy image'', ``This is a snowy image'', and ``This is a hazy image'', the model suffers from performance drops, indicating the effectiveness of learnable prompts. For the integration strategy, the use of cross-attention results in a decrease in all metrics compared to our prompt adapter. For the embedding positions, the best performance is achieved when degradation-aware features are embedded in each encoder layer of the residual estimation network.

\begin{table}[htbp]
	\centering
	\footnotesize
	\caption{Ablation study of weather-specific prompt guidance module.}
	\label{Tab:ablation2}
	\begin{threeparttable}
		\centering
		\setlength{\tabcolsep}{5mm}{
			\begin{tabular}{ccc}
				\toprule
		    	\multirow{1}{*}{Method}&
				\multirow{1}{*}{PSNR $\uparrow$}&
				\multirow{1}{*}{SSIM $\uparrow$} \cr
                    \midrule
                    Predefined Text Prompts & 32.83 & 0.9429 \cr
                    Learnable Weather Prompts & \textbf{33.01} & \textbf{0.9451} \cr
				\midrule
				Prompt Adapter & \textbf{33.01} & \textbf{0.9451} \cr
				Cross-Attention & 32.84 & 0.9419 \cr
                    \midrule
				  Encoders & \textbf{33.01} & \textbf{0.9451}\cr
                    Bottleneck & 32.57 & 0.9342\cr
                    Decoders & 32.61 & 0.9353\cr
				\bottomrule
			\end{tabular}
		}
	\end{threeparttable}
\end{table}

\begin{table}[htbp]
	\centering
	\footnotesize
	\caption{Ablation study of dynamic expert selection modulator.}
	\label{Tab:ablation3}
	\begin{threeparttable}
		\centering
		\setlength{\tabcolsep}{7mm}{
			\begin{tabular}{ccc}
				\toprule
		    	\multirow{1}{*}{Method}&
				\multirow{1}{*}{PSNR $\uparrow$}&
				\multirow{1}{*}{SSIM $\uparrow$} \cr
                    \midrule
                    \textbf{n=4} & 33.01 & 0.9451\cr
                    n=8 & 33.04 & 0.9462\cr
                    n=16 & 33.39 & 0.9483\cr
                    \midrule
                    P=0.3  & 32.85 & 0.9377\cr
                    \textbf{P=0.4}  & 33.01 & 0.9451 \cr
                    P=0.5  & 32.46 & 0.9310\cr
                    \midrule
                    $\lambda$=0.1 & 32.95 & 0.9418\cr
                    \textbf{$\lambda$=0.01} & 33.01 & 0.9451 \cr
                    $\lambda$=0.001 & 32.91 & 0.9418\cr
				\bottomrule
			\end{tabular}
		}
	\end{threeparttable}
\end{table}

\textbf{Effectiveness of DESM.} The DESM uses a dynamic weather-aware router to adaptively activate varying numbers of restoration experts for each input, enabling the model's adaptiveness to complex weather restoration. We perform ablation experiments to examine the impacts of different settings in DESM, including the total number of experts $n$, threshold $P$ in the dynamic routing mechanism, and loss weight $\lambda$. As exhibited in Table \ref{Tab:ablation3}, the performance improves with an increase in the total number of candidate experts. Yet, a larger number of experts leads to higher computational costs and redundancy. Additionally, the performance improvement is marginal when the total number of experts is set from 4 to 8. To balance the performance and efficiency, we ultimately set the total number of experts to 4. Furthermore, the model achieves the optimal performance when P is set to 0.4. When P exceeds or falls below 0.4, performance deteriorates. Similarly, setting $\lambda$ to 0.01 yields the best restoration performance.

\section{Conclusion}
\label{section VI}
In this work, we propose DA$^{2}$Diff - an innovative diffusion paradigm that learns degradation-aware adaptive priors for all-in-one weather restoration. From a new perspective, we explore the potential of large-scale vision-language model CLIP to perceive distinctive degradation characteristics via a set of learnable weather prompts. By narrowing the disparity between the degraded images and their corresponding weather-specific learnable prompt in the CLIP latent space, each learnable prompt contributes to the different weather degradation characteristics. The learned weather prompts in CLIP are incorporated into the diffusion model via the designed weather-specific prompt guidance (WPG) module, enabling the model to effectively restore multiple weather degradations. Furthermore, we propose a dynamic expert selection modulator (DESM) that flexibly assigns varying numbers of restoration experts for every input based on a dynamic weather-aware router, allowing the diffusion model to adaptively restore complex degradations in real-world scenarios. Extensive experiments on both synthetic and real-world datasets validate the superiority and effectiveness of DA$^{2}$Diff.

\vspace{5pt}
\noindent\textbf{Acknowledgements.}
The authors thank the editors and anonymous reviewers for their careful reading and valuable comments. 

\bibliographystyle{IEEEtran}
\bibliography{reference}

\begin{thebibliography}{10}
\providecommand{\url}[1]{#1}
\csname url@samestyle\endcsname
\providecommand{\newblock}{\relax}
\providecommand{\bibinfo}[2]{#2}
\providecommand{\BIBentrySTDinterwordspacing}{\spaceskip=0pt\relax}
\providecommand{\BIBentryALTinterwordstretchfactor}{4}
\providecommand{\BIBentryALTinterwordspacing}{\spaceskip=\fontdimen2\font plus
\BIBentryALTinterwordstretchfactor\fontdimen3\font minus \fontdimen4\font\relax}
\providecommand{\BIBforeignlanguage}[2]{{%
\expandafter\ifx\csname l@#1\endcsname\relax
\typeout{** WARNING: IEEEtran.bst: No hyphenation pattern has been}%
\typeout{** loaded for the language `#1'. Using the pattern for}%
\typeout{** the default language instead.}%
\else
\language=\csname l@#1\endcsname
\fi
#2}}
\providecommand{\BIBdecl}{\relax}
\BIBdecl

\bibitem{dong2024gmtnet}
C.~Dong, C.~Wang, Y.~Zhai, Y.~Li, J.~Zhou, P.~Coscia, A.~Genovese, V.~Piuri, and F.~Scotti, ``Gmtnet: Dense object detection via global dynamically matching transformer network,'' \emph{IEEE Transactions on Circuits and Systems for Video Technology}, 2024.

\bibitem{zhu2023intermediate}
Y.~Zhu, Y.~Liu, C.~Wang, S.~Wang, and M.~Lu, ``Intermediate domain based meta learning framework for adaptive object detection,'' \emph{IEEE Transactions on Circuits and Systems for Video Technology}, 2023.

\bibitem{cong2024end}
R.~Cong, H.~Sheng, D.~Yang, D.~Yang, R.~Chen, S.~Wang, and Z.~Cui, ``End-to-end semantic segmentation utilizing multi-scale baseline light field,'' \emph{IEEE Transactions on Circuits and Systems for Video Technology}, 2024.

\bibitem{wang2023dcfp}
Z.~Wang, H.~Xie, Y.~Wang, H.~Xu, and G.~Jin, ``Dcfp: Distribution calibrated filter pruning for lightweight and accurate long-tail semantic segmentation,'' \emph{IEEE Transactions on Circuits and Systems for Video Technology}, 2023.

\bibitem{he2010single}
K.~He, J.~Sun, and X.~Tang, ``Single image haze removal using dark channel prior,'' \emph{{IEEE} Trans. Pattern Anal. Mach. Intell.}, vol.~33, no.~12, pp. 2341--2353, Sep. 2010.

\bibitem{cai2016dehazenet}
B.~Cai, X.~Xu, K.~Jia, C.~Qing, and D.~Tao, ``Dehazenet: An end-to-end system for single image haze removal,'' \emph{IEEE transactions on image processing}, vol.~25, no.~11, pp. 5187--5198, 2016.

\bibitem{liu2019griddehazenet}
X.~Liu, Y.~Ma, Z.~Shi, and J.~Chen, ``Griddehazenet: Attention-based multi-scale network for image dehazing,'' in \emph{Proceedings of the IEEE/CVF international conference on computer vision}, 2019, pp. 7314--7323.

\bibitem{song2022wsamf}
X.~Song, D.~Zhou, W.~Li, H.~Ding, Y.~Dai, and L.~Zhang, ``Wsamf-net: Wavelet spatial attention-based multistream feedback network for single image dehazing,'' \emph{IEEE Transactions on Circuits and Systems for Video Technology}, vol.~33, no.~2, pp. 575--588, 2022.

\bibitem{wang2024uncertainty}
B.~Wang, Q.~Ning, F.~Wu, X.~Li, W.~Dong, and G.~Shi, ``Uncertainty modeling of the transmission map for single image dehazing,'' \emph{IEEE Transactions on Circuits and Systems for Video Technology}, 2024.

\bibitem{li2018recurrent}
X.~Li, J.~Wu, Z.~Lin, H.~Liu, and H.~Zha, ``Recurrent squeeze-and-excitation context aggregation net for single image deraining,'' in \emph{Proceedings of the European conference on computer vision (ECCV)}, 2018, pp. 254--269.

\bibitem{qian2018attentive}
R.~Qian, R.~T. Tan, W.~Yang, J.~Su, and J.~Liu, ``Attentive generative adversarial network for raindrop removal from a single image,'' in \emph{Proceedings of the IEEE conference on computer vision and pattern recognition}, 2018, pp. 2482--2491.

\bibitem{wang2019spatial}
T.~Wang, X.~Yang, K.~Xu, S.~Chen, Q.~Zhang, and R.~W. Lau, ``Spatial attentive single-image deraining with a high quality real rain dataset,'' in \emph{Proceedings of the IEEE/CVF conference on computer vision and pattern recognition}, 2019, pp. 12\,270--12\,279.

\bibitem{du2020conditional}
Y.~Du, J.~Xu, X.~Zhen, M.-M. Cheng, and L.~Shao, ``Conditional variational image deraining,'' \emph{IEEE Transactions on Image Processing}, vol.~29, pp. 6288--6301, 2020.

\bibitem{cai2022multiscale}
L.~Cai, Y.~Fu, W.~Huo, Y.~Xiang, T.~Zhu, Y.~Zhang, H.~Zeng, and D.~Zeng, ``Multiscale attentive image de-raining networks via neural architecture search,'' \emph{IEEE Transactions on Circuits and Systems for Video Technology}, vol.~33, no.~2, pp. 618--633, 2022.

\bibitem{yan2024glgfn}
T.~Yan, X.~Zhu, X.~Chen, W.~He, C.~Wang, Y.~Yang, Y.~Wang, and X.~Chang, ``Glgfn: Global-local grafting fusion network for high-resolution image deraining,'' \emph{IEEE Transactions on Circuits and Systems for Video Technology}, 2024.

\bibitem{liu2018desnownet}
Y.-F. Liu, D.-W. Jaw, S.-C. Huang, and J.-N. Hwang, ``Desnownet: Context-aware deep network for snow removal,'' \emph{IEEE Transactions on Image Processing}, vol.~27, no.~6, pp. 3064--3073, 2018.

\bibitem{chen2020jstasr}
W.-T. Chen, H.-Y. Fang, J.-J. Ding, C.-C. Tsai, and S.-Y. Kuo, ``Jstasr: Joint size and transparency-aware snow removal algorithm based on modified partial convolution and veiling effect removal,'' in \emph{Computer Vision--ECCV 2020: 16th European Conference, Glasgow, UK, August 23--28, 2020, Proceedings, Part XXI 16}.\hskip 1em plus 0.5em minus 0.4em\relax Springer, 2020, pp. 754--770.

\bibitem{jaw2020desnowgan}
D.-W. Jaw, S.-C. Huang, and S.-Y. Kuo, ``Desnowgan: An efficient single image snow removal framework using cross-resolution lateral connection and gans,'' \emph{IEEE Transactions on Circuits and Systems for Video Technology}, vol.~31, no.~4, pp. 1342--1350, 2020.

\bibitem{zhang2021deep}
K.~Zhang, R.~Li, Y.~Yu, W.~Luo, and C.~Li, ``Deep dense multi-scale network for snow removal using semantic and depth priors,'' \emph{IEEE Transactions on Image Processing}, vol.~30, pp. 7419--7431, 2021.

\bibitem{zamir2021multi}
S.~W. Zamir, A.~Arora, S.~Khan, M.~Hayat, F.~S. Khan, M.-H. Yang, and L.~Shao, ``Multi-stage progressive image restoration,'' in \emph{Proceedings of the IEEE/CVF conference on computer vision and pattern recognition}, 2021, pp. 14\,821--14\,831.

\bibitem{chen2022simple}
L.~Chen, X.~Chu, X.~Zhang, and J.~Sun, ``Simple baselines for image restoration,'' in \emph{European conference on computer vision}.\hskip 1em plus 0.5em minus 0.4em\relax Springer, 2022, pp. 17--33.

\bibitem{zamir2022restormer}
S.~W. Zamir, A.~Arora, S.~Khan, M.~Hayat, F.~S. Khan, and M.-H. Yang, ``Restormer: Efficient transformer for high-resolution image restoration,'' in \emph{Proceedings of the IEEE/CVF conference on computer vision and pattern recognition}, 2022, pp. 5728--5739.

\bibitem{tu2022maxim}
Z.~Tu, H.~Talebi, H.~Zhang, F.~Yang, P.~Milanfar, A.~Bovik, and Y.~Li, ``Maxim: Multi-axis mlp for image processing,'' in \emph{Proceedings of the IEEE/CVF conference on computer vision and pattern recognition}, 2022, pp. 5769--5780.

\bibitem{li2020all}
R.~Li, R.~T. Tan, and L.-F. Cheong, ``All in one bad weather removal using architectural search,'' in \emph{Proceedings of the IEEE/CVF conference on computer vision and pattern recognition}, 2020, pp. 3175--3185.

\bibitem{chen2022learning}
W.-T. Chen, Z.-K. Huang, C.-C. Tsai, H.-H. Yang, J.-J. Ding, and S.-Y. Kuo, ``Learning multiple adverse weather removal via two-stage knowledge learning and multi-contrastive regularization: Toward a unified model,'' in \emph{Proceedings of the IEEE/CVF Conference on Computer Vision and Pattern Recognition}, 2022, pp. 17\,653--17\,662.

\bibitem{valanarasu2022transweather}
J.~M.~J. Valanarasu, R.~Yasarla, and V.~M. Patel, ``Transweather: Transformer-based restoration of images degraded by adverse weather conditions,'' in \emph{Proceedings of the IEEE/CVF Conference on Computer Vision and Pattern Recognition}, 2022, pp. 2353--2363.

\bibitem{zhu2023learning}
Y.~Zhu, T.~Wang, X.~Fu, X.~Yang, X.~Guo, J.~Dai, Y.~Qiao, and X.~Hu, ``Learning weather-general and weather-specific features for image restoration under multiple adverse weather conditions,'' in \emph{Proceedings of the IEEE/CVF conference on computer vision and pattern recognition}, 2023, pp. 21\,747--21\,758.

\bibitem{ozdenizci2023restoring}
O.~{\"O}zdenizci and R.~Legenstein, ``Restoring vision in adverse weather conditions with patch-based denoising diffusion models,'' \emph{IEEE Transactions on Pattern Analysis and Machine Intelligence}, vol.~45, no.~8, pp. 10\,346--10\,357, 2023.

\bibitem{gao2023frequency}
T.~Gao, Y.~Wen, K.~Zhang, J.~Zhang, T.~Chen, L.~Liu, and W.~Luo, ``Frequency-oriented efficient transformer for all-in-one weather-degraded image restoration,'' \emph{IEEE Transactions on Circuits and Systems for Video Technology}, 2023.

\bibitem{zhu2024mwformer}
R.~Zhu, Z.~Tu, J.~Liu, A.~C. Bovik, and Y.~Fan, ``Mwformer: Multi-weather image restoration using degradation-aware transformers,'' \emph{IEEE Transactions on Image Processing}, 2024.

\bibitem{li2024multi}
C.~Li, F.~Sun, H.~Zhou, Y.~Xie, Z.~Li, and L.~Zhu, ``Multi-weather restoration: An efficient prompt-guided convolution architecture,'' \emph{IEEE Transactions on Circuits and Systems for Video Technology}, 2024.

\bibitem{ye2023adverse}
T.~Ye, S.~Chen, J.~Bai, J.~Shi, C.~Xue, J.~Jiang, J.~Yin, E.~Chen, and Y.~Liu, ``Adverse weather removal with codebook priors,'' in \emph{Proceedings of the IEEE/CVF International Conference on Computer Vision}, 2023, pp. 12\,653--12\,664.

\bibitem{ho2020denoising}
J.~Ho, A.~Jain, and P.~Abbeel, ``Denoising diffusion probabilistic models,'' \emph{Advances in neural information processing systems}, vol.~33, pp. 6840--6851, 2020.

\bibitem{saharia2022image}
C.~Saharia, J.~Ho, W.~Chan, T.~Salimans, D.~J. Fleet, and M.~Norouzi, ``Image super-resolution via iterative refinement,'' \emph{IEEE Transactions on Pattern Analysis and Machine Intelligence}, vol.~45, no.~4, pp. 4713--4726, 2022.

\bibitem{radford2021learning}
A.~Radford, J.~W. Kim, C.~Hallacy, A.~Ramesh, G.~Goh, S.~Agarwal, G.~Sastry, A.~Askell, P.~Mishkin, J.~Clark \emph{et~al.}, ``Learning transferable visual models from natural language supervision,'' in \emph{International conference on machine learning}.\hskip 1em plus 0.5em minus 0.4em\relax PMLR, 2021, pp. 8748--8763.

\bibitem{zheng2024selective}
D.~Zheng, X.-M. Wu, S.~Yang, J.~Zhang, J.-F. Hu, and W.-S. Zheng, ``Selective hourglass mapping for universal image restoration based on diffusion model,'' in \emph{Proceedings of the IEEE/CVF Conference on Computer Vision and Pattern Recognition}, 2024, pp. 25\,445--25\,455.

\bibitem{luo2023controlling}
Z.~Luo, F.~K. Gustafsson, Z.~Zhao, J.~Sj{\"o}lund, and T.~B. Sch{\"o}n, ``Controlling vision-language models for universal image restoration,'' \emph{arXiv preprint arXiv:2310.01018}, vol.~3, no.~8, 2023.

\bibitem{jiang2025autodir}
Y.~Jiang, Z.~Zhang, T.~Xue, and J.~Gu, ``Autodir: Automatic all-in-one image restoration with latent diffusion,'' in \emph{European Conference on Computer Vision}.\hskip 1em plus 0.5em minus 0.4em\relax Springer, 2025, pp. 340--359.

\bibitem{shazeer2017outrageously}
N.~Shazeer, A.~Mirhoseini, K.~Maziarz, A.~Davis, Q.~Le, G.~Hinton, and J.~Dean, ``Outrageously large neural networks: The sparsely-gated mixture-of-experts layer,'' \emph{arXiv preprint arXiv:1701.06538}, 2017.

\bibitem{li2017aod}
B.~Li, X.~Peng, Z.~Wang, J.~Xu, and D.~Feng, ``Aod-net: All-in-one dehazing network,'' in \emph{Proc. IEEE Int. Conf. Comput. Vis. (ICCV)}, Oct. 2017, pp. 4770--4778.

\bibitem{zhang2018densely}
H.~Zhang and V.~M. Patel, ``Densely connected pyramid dehazing network,'' in \emph{Proceedings of the IEEE conference on computer vision and pattern recognition}, 2018, pp. 3194--3203.

\bibitem{narasimhan2000chromatic}
S.~G. Narasimhan and S.~K. Nayar, ``Chromatic framework for vision in bad weather,'' in \emph{Proc. IEEE/CVF Conf. Comput. Vis. Pattern Recognit. (CVPR)}, vol.~1, Jun. 2000, pp. 598--605.

\bibitem{qu2019enhanced}
Y.~Qu, Y.~Chen, J.~Huang, and Y.~Xie, ``Enhanced pix2pix dehazing network,'' in \emph{Proceedings of the IEEE/CVF conference on computer vision and pattern recognition}, 2019, pp. 8160--8168.

\bibitem{song2023vision}
Y.~Song, Z.~He, H.~Qian, and X.~Du, ``Vision transformers for single image dehazing,'' \emph{IEEE Transactions on Image Processing}, vol.~32, pp. 1927--1941, 2023.

\bibitem{quan2019deep}
Y.~Quan, S.~Deng, Y.~Chen, and H.~Ji, ``Deep learning for seeing through window with raindrops,'' in \emph{Proceedings of the IEEE/CVF International Conference on Computer Vision}, 2019, pp. 2463--2471.

\bibitem{patashnik2021styleclip}
O.~Patashnik, Z.~Wu, E.~Shechtman, D.~Cohen-Or, and D.~Lischinski, ``Styleclip: Text-driven manipulation of stylegan imagery,'' in \emph{Proceedings of the IEEE/CVF international conference on computer vision}, 2021, pp. 2085--2094.

\bibitem{wei2022hairclip}
T.~Wei, D.~Chen, W.~Zhou, J.~Liao, Z.~Tan, L.~Yuan, W.~Zhang, and N.~Yu, ``Hairclip: Design your hair by text and reference image,'' in \emph{Proceedings of the IEEE/CVF Conference on Computer Vision and Pattern Recognition}, 2022, pp. 18\,072--18\,081.

\bibitem{crowson2022vqgan}
K.~Crowson, S.~Biderman, D.~Kornis, D.~Stander, E.~Hallahan, L.~Castricato, and E.~Raff, ``Vqgan-clip: Open domain image generation and editing with natural language guidance,'' in \emph{European Conference on Computer Vision}.\hskip 1em plus 0.5em minus 0.4em\relax Springer, 2022, pp. 88--105.

\bibitem{wang2022cris}
Z.~Wang, Y.~Lu, Q.~Li, X.~Tao, Y.~Guo, M.~Gong, and T.~Liu, ``Cris: Clip-driven referring image segmentation,'' in \emph{Proceedings of the IEEE/CVF conference on computer vision and pattern recognition}, 2022, pp. 11\,686--11\,695.

\bibitem{rao2022denseclip}
Y.~Rao, W.~Zhao, G.~Chen, Y.~Tang, Z.~Zhu, G.~Huang, J.~Zhou, and J.~Lu, ``Denseclip: Language-guided dense prediction with context-aware prompting,'' in \emph{Proceedings of the IEEE/CVF conference on computer vision and pattern recognition}, 2022, pp. 18\,082--18\,091.

\bibitem{liang2023iterative}
Z.~Liang, C.~Li, S.~Zhou, R.~Feng, and C.~C. Loy, ``Iterative prompt learning for unsupervised backlit image enhancement,'' in \emph{Proceedings of the IEEE/CVF International Conference on Computer Vision}, 2023, pp. 8094--8103.

\bibitem{sun2024coser}
H.~Sun, W.~Li, J.~Liu, H.~Chen, R.~Pei, X.~Zou, Y.~Yan, and Y.~Yang, ``Coser: Bridging image and language for cognitive super-resolution,'' in \emph{Proceedings of the IEEE/CVF Conference on Computer Vision and Pattern Recognition}, 2024, pp. 25\,868--25\,878.

\bibitem{sener2018multi}
O.~Sener and V.~Koltun, ``Multi-task learning as multi-objective optimization,'' \emph{Advances in neural information processing systems}, vol.~31, 2018.

\bibitem{jacobs1991adaptive}
R.~A. Jacobs, M.~I. Jordan, S.~J. Nowlan, and G.~E. Hinton, ``Adaptive mixtures of local experts,'' \emph{Neural computation}, vol.~3, no.~1, pp. 79--87, 1991.

\bibitem{du2022glam}
N.~Du, Y.~Huang, A.~M. Dai, S.~Tong, D.~Lepikhin, Y.~Xu, M.~Krikun, Y.~Zhou, A.~W. Yu, O.~Firat \emph{et~al.}, ``Glam: Efficient scaling of language models with mixture-of-experts,'' in \emph{International Conference on Machine Learning}.\hskip 1em plus 0.5em minus 0.4em\relax PMLR, 2022, pp. 5547--5569.

\bibitem{enzweiler2011multilevel}
M.~Enzweiler and D.~M. Gavrila, ``A multilevel mixture-of-experts framework for pedestrian classification,'' \emph{IEEE Transactions on Image Processing}, vol.~20, no.~10, pp. 2967--2979, 2011.

\bibitem{riquelme2021scaling}
C.~Riquelme, J.~Puigcerver, B.~Mustafa, M.~Neumann, R.~Jenatton, A.~Susano~Pinto, D.~Keysers, and N.~Houlsby, ``Scaling vision with sparse mixture of experts,'' \emph{Advances in Neural Information Processing Systems}, vol.~34, pp. 8583--8595, 2021.

\bibitem{ng2023botbuster}
L.~H.~X. Ng and K.~M. Carley, ``Botbuster: Multi-platform bot detection using a mixture of experts,'' in \emph{Proceedings of the international AAAI conference on web and social media}, vol.~17, 2023, pp. 686--697.

\bibitem{yang2024multi}
Y.~Yang, P.-T. Jiang, Q.~Hou, H.~Zhang, J.~Chen, and B.~Li, ``Multi-task dense prediction via mixture of low-rank experts,'' in \emph{Proceedings of the IEEE/CVF Conference on Computer Vision and Pattern Recognition}, 2024, pp. 27\,927--27\,937.

\bibitem{zhang2024efficient}
R.~Zhang, Y.~Luo, J.~Liu, H.~Yang, Z.~Dong, D.~Gudovskiy, T.~Okuno, Y.~Nakata, K.~Keutzer, Y.~Du \emph{et~al.}, ``Efficient deweahter mixture-of-experts with uncertainty-aware feature-wise linear modulation,'' in \emph{Proceedings of the AAAI Conference on Artificial Intelligence}, vol.~38, no.~15, 2024, pp. 16\,812--16\,820.

\bibitem{kingma2013auto}
D.~P. Kingma, ``Auto-encoding variational bayes,'' \emph{arXiv preprint arXiv:1312.6114}, 2013.

\bibitem{kingma2019introduction}
D.~P. Kingma, M.~Welling \emph{et~al.}, ``An introduction to variational autoencoders,'' \emph{Foundations and Trends in Machine Learning}, vol.~12, no.~4, pp. 307--392, 2019.

\bibitem{song2020denoising}
J.~Song, C.~Meng, and S.~Ermon, ``Denoising diffusion implicit models,'' \emph{arXiv preprint arXiv:2010.02502}, 2020.

\bibitem{fedus2022switch}
W.~Fedus, B.~Zoph, and N.~Shazeer, ``Switch transformers: Scaling to trillion parameter models with simple and efficient sparsity,'' \emph{Journal of Machine Learning Research}, vol.~23, no. 120, pp. 1--39, 2022.

\bibitem{li2019heavy}
R.~Li, L.-F. Cheong, and R.~T. Tan, ``Heavy rain image restoration: Integrating physics model and conditional adversarial learning,'' in \emph{Proceedings of the IEEE/CVF conference on computer vision and pattern recognition}, 2019, pp. 1633--1642.

\bibitem{quan2021removing}
R.~Quan, X.~Yu, Y.~Liang, and Y.~Yang, ``Removing raindrops and rain streaks in one go,'' in \emph{Proceedings of the IEEE/CVF conference on computer vision and pattern recognition}, 2021, pp. 9147--9156.

\bibitem{paszke2019pytorch}
A.~Paszke, S.~Gross, F.~Massa, A.~Lerer, J.~Bradbury, G.~Chanan, T.~Killeen, Z.~Lin, N.~Gimelshein, L.~Antiga \emph{et~al.}, ``Pytorch: An imperative style, high-performance deep learning library,'' \emph{Advances in neural information processing systems}, vol.~32, 2019.

\bibitem{song2020improved}
Y.~Song and S.~Ermon, ``Improved techniques for training score-based generative models,'' \emph{Advances in neural information processing systems}, vol.~33, pp. 12\,438--12\,448, 2020.

\bibitem{assessment2004error}
I.~Q. Assessment, ``From error visibility to structural similarity,'' \emph{IEEE transactions on image processing}, vol.~13, no.~4, p.~93, 2004.

\bibitem{zhang2015feature}
L.~Zhang, L.~Zhang, and A.~C. Bovik, ``A feature-enriched completely blind image quality evaluator,'' \emph{IEEE Transactions on Image Processing}, vol.~24, no.~8, pp. 2579--2591, 2015.

\bibitem{wang2023exploring}
J.~Wang, K.~C. Chan, and C.~C. Loy, ``Exploring clip for assessing the look and feel of images,'' in \emph{Proceedings of the AAAI Conference on Artificial Intelligence}, vol.~37, no.~2, 2023, pp. 2555--2563.

\bibitem{yang2022maniqa}
S.~Yang, T.~Wu, S.~Shi, S.~Lao, Y.~Gong, M.~Cao, J.~Wang, and Y.~Yang, ``Maniqa: Multi-dimension attention network for no-reference image quality assessment,'' in \emph{Proceedings of the IEEE/CVF Conference on Computer Vision and Pattern Recognition}, 2022, pp. 1191--1200.

\bibitem{isola2017image}
P.~Isola, J.-Y. Zhu, T.~Zhou, and A.~A. Efros, ``Image-to-image translation with conditional adversarial networks,'' in \emph{Proceedings of the IEEE conference on computer vision and pattern recognition}, 2017, pp. 1125--1134.

\bibitem{liu2019dual}
X.~Liu, M.~Suganuma, Z.~Sun, and T.~Okatani, ``Dual residual networks leveraging the potential of paired operations for image restoration,'' in \emph{Proceedings of the IEEE/CVF Conference on Computer Vision and Pattern Recognition}, 2019, pp. 7007--7016.

\bibitem{xiao2022image}
J.~Xiao, X.~Fu, A.~Liu, F.~Wu, and Z.-J. Zha, ``Image de-raining transformer,'' \emph{IEEE Transactions on Pattern Analysis and Machine Intelligence}, vol.~45, no.~11, pp. 12\,978--12\,995, 2022.

\bibitem{chen2023sparse}
S.~Chen, T.~Ye, J.~Bai, E.~Chen, J.~Shi, and L.~Zhu, ``Sparse sampling transformer with uncertainty-driven ranking for unified removal of raindrops and rain streaks,'' in \emph{Proceedings of the IEEE/CVF International Conference on Computer Vision}, 2023, pp. 13\,106--13\,117.

\bibitem{zhu2017unpaired}
J.-Y. Zhu, T.~Park, P.~Isola, and A.~A. Efros, ``Unpaired image-to-image translation using cycle-consistent adversarial networks,'' in \emph{Proceedings of the IEEE international conference on computer vision}, 2017, pp. 2223--2232.

\bibitem{jiang2021rain}
K.~Jiang, Z.~Wang, P.~Yi, C.~Chen, Z.~Wang, X.~Wang, J.~Jiang, and C.-W. Lin, ``Rain-free and residue hand-in-hand: A progressive coupled network for real-time image deraining,'' \emph{IEEE Transactions on Image Processing}, vol.~30, pp. 7404--7418, 2021.

\end{thebibliography}

\vfill
\end{document}